\documentclass[]{template}

\usepackage[utf8]{inputenc}
\usepackage[T1]{fontenc}
\usepackage{url}
\usepackage{booktabs}
\usepackage{multirow}
\usepackage{makecell}
\usepackage{array}
\usepackage{colortbl}
\usepackage{multicol}
\usepackage{amsfonts}
\usepackage{nicefrac}
\usepackage{microtype}
\usepackage[dvipsnames,table]{xcolor}
\usepackage{latexsym}
\usepackage{graphicx}
\usepackage{float}
\usepackage{subcaption}
\usepackage{wrapfig}
\usepackage{bm}
\usepackage{tabularx}
\usepackage{ragged2e}
\newcolumntype{L}{>{\RaggedRight\hangafter=1\hangindent=0em}X}
\usepackage{enumitem}
\usepackage{amsmath}
\usepackage{amssymb}
\usepackage{mathtools}
\usepackage{amsthm}
\usepackage{arydshln}
\usepackage{pifont}
\usepackage{marvosym}
\usepackage{placeins}
\usepackage{xspace}
\usepackage[most]{tcolorbox}

\setboolean{logo}{true}

\definecolor{myred}{HTML}{C00000}
\definecolor{mygreen}{HTML}{00B050}
\definecolor{myblue}{HTML}{0070C0}
\definecolor{myyellow}{HTML}{FFC000}

\hypersetup{
    colorlinks=true,
    linkcolor=red,
    citecolor=Cerulean,
    filecolor=magenta,
    urlcolor=magenta,
}

\usepackage[capitalize,noabbrev]{cleveref}
\crefname{figure}{Fig.}{Figs.}
\Crefname{figure}{Fig.}{Figs.}
\crefname{table}{Tab.}{Tabs.}
\Crefname{table}{Tab.}{Tabs.}
\crefname{section}{Sec.}{Secs.}
\Crefname{section}{Sec.}{Secs.}

\theoremstyle{plain}

\theoremstyle{definition}

\theoremstyle{remark}

\DeclareCaptionLabelFormat{cont}{#1~#2\alph{ContinuedFloat}}
\tcbset{
  promptbox/.style={
    top=10pt,
    colback=lightgray!20,
    colframe=Black,
    colbacktitle=NavyBlue,
    enhanced,
    center,
    attach boxed title to top center={yshift=-0.1in,xshift=0.0in},
    boxed title style={boxrule=0pt,colframe=white,},
  },
  takeawaybox/.style={
    top=10pt,
    colback=lightgray!20,
    colframe=Black,
    colbacktitle=BurntOrange,
    enhanced,
    center,
    attach boxed title to top center={yshift=-0.1in,xshift=0.0in},
    boxed title style={boxrule=0pt,colframe=white,},
  },
  observationbox/.style={
    top=10pt,
    colback=lightgray!20,
    colframe=Black,
    colbacktitle=YellowGreen,
    enhanced,
    center,
    attach boxed title to top center={yshift=-0.1in,xshift=0.0in},
    boxed title style={boxrule=0pt,colframe=white,},
  }
}
\newtcolorbox{promptbox}[2][]{promptbox, title=#2,#1}
\newtcolorbox{takeawaybox}[2][]{takeawaybox, title=#2,#1}
\newtcolorbox{observationbox}[2][]{observationbox, title=#2,#1}

\newcommand\blfootnote[1]{%
  \begingroup
  \renewcommand\thefootnote{}\footnote{#1}%
  \addtocounter{footnote}{-1}%
  \endgroup
}

\title{CapRL++: Unified Reinforcement Learning with Verifiable Rewards for Dense Image and Video Captioning}

\author[1]{Penghui Yang$^*$}
\author[2]{Long Xing$^*$}
\author[3]{Xiaoyi Dong}
\author[4]{Yuhang Zang$^\dagger$}
\author[4]{Yuhang Cao}
\author[5]{Yibin Wang}
\author[4]{Yujie Zhou}
\author[4]{Jiazi Bu}
\author[4]{Jianze Liang}
\author[6]{Qidong Huang}
\author[5]{Jiaqi Wang}
\author[2]{Feng Wu$^\dagger$}
\author[7]{Dahua Lin}

\affil[1]{Tsinghua University}
\affil[2]{University of Science and Technology of China}
\affil[3]{Microsoft}
\affil[4]{Shanghai AI Laboratory}
\affil[5]{Shanghai Innovation Institute}
\affil[6]{Alibaba Cloud}
\affil[7]{The Chinese University of Hong Kong}

\begin{abstract}
Image and video captioning are fundamental tasks that bridge the visual and linguistic domains, 
playing a critical role in pre-training Large Vision-Language Models (LVLMs). 
Current state-of-the-art captioning models are typically trained with Supervised Fine-Tuning (SFT), 
a paradigm that relies on expensive, non-scalable annotations and often causes models to memorize specific ground-truth answers, limiting their generality and
ability to generate diverse, creative descriptions.
To overcome these limitations, we propose applying the Reinforcement Learning with Verifiable Rewards (RLVR) paradigm to the open-ended task of multimodal captioning. 
We introduce Captioning Reinforcement Learning++ (CapRL++), a novel reference-free training framework that redefines caption quality through its utility: 
a high-quality caption should enable a non-visual language model to accurately answer questions about the corresponding visual content. 
CapRL++ employs a decoupled two-stage pipeline where an LVLM generates a caption, and the objective reward is derived from the accuracy of a separate, vision-free LLM answering Multiple-Choice Questions based solely on that caption.
Evaluations on more than 20 image and video benchmarks show that CapRL++ improves dense caption quality and strengthens caption-based pretraining across various tasks such as spatial and temporal understanding.
Pretraining on our scalable image and video caption datasets annotated by CapRL++ yields substantial downstream gains. 
Furthermore, within the Prism Framework for caption quality evaluation, compact models trained with CapRL++ achieve dense captioning performance comparable to, 
and in some cases surpassing, substantially larger models (e.g., Qwen2.5-VL-72B and Qwen3-VL-235B-A22B). 
Results validate that CapRL++ effectively trains models to produce generalizable, high-fidelity descriptions, establishing a robust foundation beyond the limitations of traditional SFT.
\end{abstract}

\begin{document}

\blfootnote{$^*$ Equal contribution. $^\dagger$ Corresponding authors: Yuhang Zang (zangyuhang@pjlab.org.cn) and Feng Wu (fengwu@ustc.edu.cn).}
\blfootnote{Code is available at \url{https://github.com/InternLM/CapRL}.}

\maketitle


\section{Introduction}
\label{sec:intro}

\begin{figure*}[t!]
    \centering
    \includegraphics[width=\textwidth]{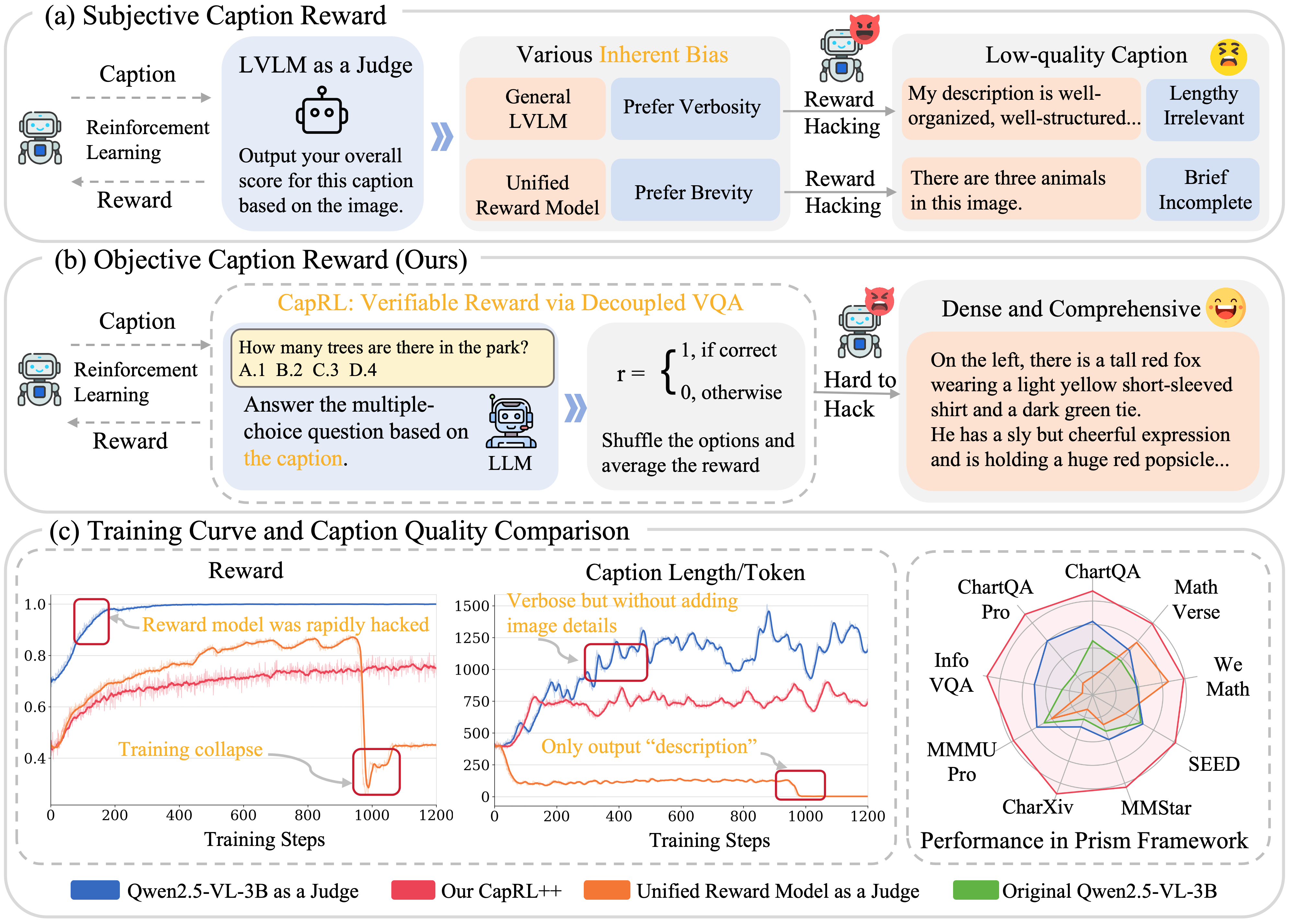}
    \vspace{-10pt}
    \caption{\textbf{(a) Existing Reward Models:} 
    Current LVLM-as-a-judge/reward models suffer from limitations like rewarding verbosity or brevity, leading to low-quality captions and reward hacking. 
    \textbf{(b) Our CapRL++:} CapRL++ uses a decoupled two-stage VQA approach to provide objective rewards for captions. The same paradigm covers both image and video captioning.
    \textbf{(c) CapRL++'s Advantage:} CapRL++ outperforms previous subjective reward methods, as shown by training curves and higher performance in the Prism evaluation setting.}
    \vspace{-15pt}
    \label{fig:teaser}
\end{figure*}

The image and video captioning tasks \cite{karpathy2015deep,vinyals2015show,venugopalan2015sequence,donahue2015long,vinyals2016show,stefanini2022show}, 
which generate natural language descriptions for a given image or video, 
bridge the visual and linguistic worlds.
The captioning capability is fundamental to a range of downstream uses, including vision-language models such as CLIP \cite{radford2021learning} that learn a shared image-text embedding space.
Captions are also a core component in the pre-training of Large Vision-Language Models (LVLMs) \cite{liu2023visual,xu2023multimodal,zhang2024vision}, where the model learns to align visual content with linguistic descriptions at scale before being fine-tuned for downstream tasks.

Given the importance of visual captioning, there is a strong need for captioning models that can provide dense and accurate descriptions for both images and videos.
Most modern captioning models \cite{chen2024sharegpt4v,rotstein2024fusecap,vasu2025fastvlm,li2025otter,li2025uni} are trained on top of LVLMs using Supervised Fine-Tuning (SFT).
While effective, SFT requires large datasets annotated by humans or proprietary models, which are \textbf{expensive} and \textbf{not scalable}.
Furthermore, visual captioning is an inherently open-ended problem, where a single image or video can be accurately described by a wide variety of captions.
Since SFT models are trained to match a single ground-truth description for each input, they tend to \textbf{memorize specific answers} rather than learn the underlying visual concepts.
As a result, SFT models become \textbf{less general} and struggle to generate the diverse range of valid captions possible for a given image or video.

The limitations of SFT have led to a recent paradigm shift in the LVLMs post-training toward Reinforcement Learning with Verifiable Rewards (RLVR) \cite{lambert2024tulu}.
RLVR trains models with clear and objective rewards from a verifier, such as a binary signal of correctness for mathematical reasoning \cite{guo2025deepseek}.
Unlike SFT, which teaches a model to mimic a single ground-truth response, RLVR encourages the model to generate more diverse and robust outputs that meet the verifiable criteria.
Our objective is to design a powerful and scalable RLVR training paradigm for image and video captioning, producing a wider variety of creative yet accurate descriptions.

However, applying RLVR to open-ended tasks like visual captioning is challenging, primarily due to the difficulty of designing an \textit{objective} reward function.
A good caption can be \textit{subjective}, with multiple valid descriptions possible for the same image or video \cite{wang2020diversity}.
As shown in \cref{fig:teaser}~\textbf{(a)}, early studies fail to provide accurate reward signals for RL training.
Using \textbf{reward models} \cite{liu2025inference,su2025crossing,lu2025writing} or \textbf{LLM-as-a-judge} \cite{gunjal2025rubrics} as feedback is vulnerable to \textit{reward hacking}, where the captioning model exploits evaluator biases (e.g., a preference for verbose or brief outputs) instead of producing a high-quality response.
Moreover, it is difficult to create effective rubrics or evaluation prompts for LVLM-as-a-judge methods because captions are free-form and encode substantial information.
Using \textbf{reference answers} as rewards \cite{gurung2025learning,yu2025rlpr} via metrics such as ROUGE \cite{lin2004rouge} and BLEU \cite{papineni2002bleu} is similarly constrained when evaluating complex and long-form captions.
\cref{fig:teaser}~\textbf{(c)} further illustrates these limitations, showing reward hacking and unstable training curves for prior subjective rewards.
This challenge is further amplified when extending the image captioning task to the \textbf{video domain}, where temporal causality and timestamp alignment introduce new dimensions of potential hallucinations and redundancy.

To design the \textit{objective} RLVR reward function for the \textit{subjective} visual captioning task, we introduce a novel perspective: a caption's quality is proportional to its utility.
When the caption is detailed and accurate, a text-based LLM that cannot directly ``see'' the image or video can still answer Visual Question Answering (VQA) questions about the source content from the caption alone.
For example, given the question ``What color is the frisbee?'', the LLM finds the phrase ``red frisbee'' in the caption and correctly answers ``red.''
Driven by this motivation, we present a decoupled two-stage pipeline dubbed \textbf{Cap}tioning \textbf{R}einforcement \textbf{L}earning++ (\textbf{CapRL++}), as shown in \cref{fig:teaser}~\textbf{(b)}.
Specifically, the reward in CapRL++ is determined by how well a caption generated by an LVLM enables a separate vision-free LLM to answer Multiple-Choice Questions (MCQs) about the source image or video, and the LLM's resulting accuracy serves as the objective reward for RLVR training.

A preliminary version of this work, CapRL \cite{xing2025caprl}, was presented at ICLR 2026. 
This journal version substantially extends the conference paper from image-only captioning to a unified image-video captioning framework. 
Compared with the preliminary version, this work introduces new spatio-temporal verifiable rewards for video captioning, proposes a Spatial-Anchored Bootstrapping (SpaBoot) recipe for image-to-video transfer, expands the QA curation and training corpus from image-only supervision to both image and video data, constructs the CapRL-Video-178K dense video caption corpus, and provides substantially broader evaluations, including video captioning, video continual pretraining, temporal grounding, cross-modal transfer, scaling behavior, and additional ablation studies. 
The main extensions and contributions of this journal version are summarized as follows.
(1) We generalize the reference-free RLVR paradigm of CapRL to both image and video captioning, where caption quality is measured by its downstream utility for a vision-free LLM to answer multiple-choice questions about the source visual content.
(2) We design spatio-temporal verifiable rewards for dense video captioning, combining MCQ-based utility, timestamp-format constraints, and length-aware regularization to improve temporal grounding and reduce redundancy.
(3) We introduce Spatial-Anchored Bootstrapping (SpaBoot), an image-to-video training recipe that first builds strong spatial perception on static images and then transfers it to dynamic videos for temporal reasoning. Our analysis further reveals meaningful cross-modal transfer between image and video supervision.
(4) We evaluate CapRL++ on more than 20 image and video benchmarks, including caption-quality evaluation, continual-pretraining evaluation, temporal grounding, cross-modal transfer, and ablations on reward design, SpaBoot strategy, data scale, and caption length. The resulting compact models achieve dense captioning performance comparable to, and in some cases surpassing, substantially larger LVLMs such as Qwen2.5-VL-72B and Qwen3-VL-235B-A22B.

\begin{figure*}[t]
    \centering
    \includegraphics[width=0.99\textwidth]{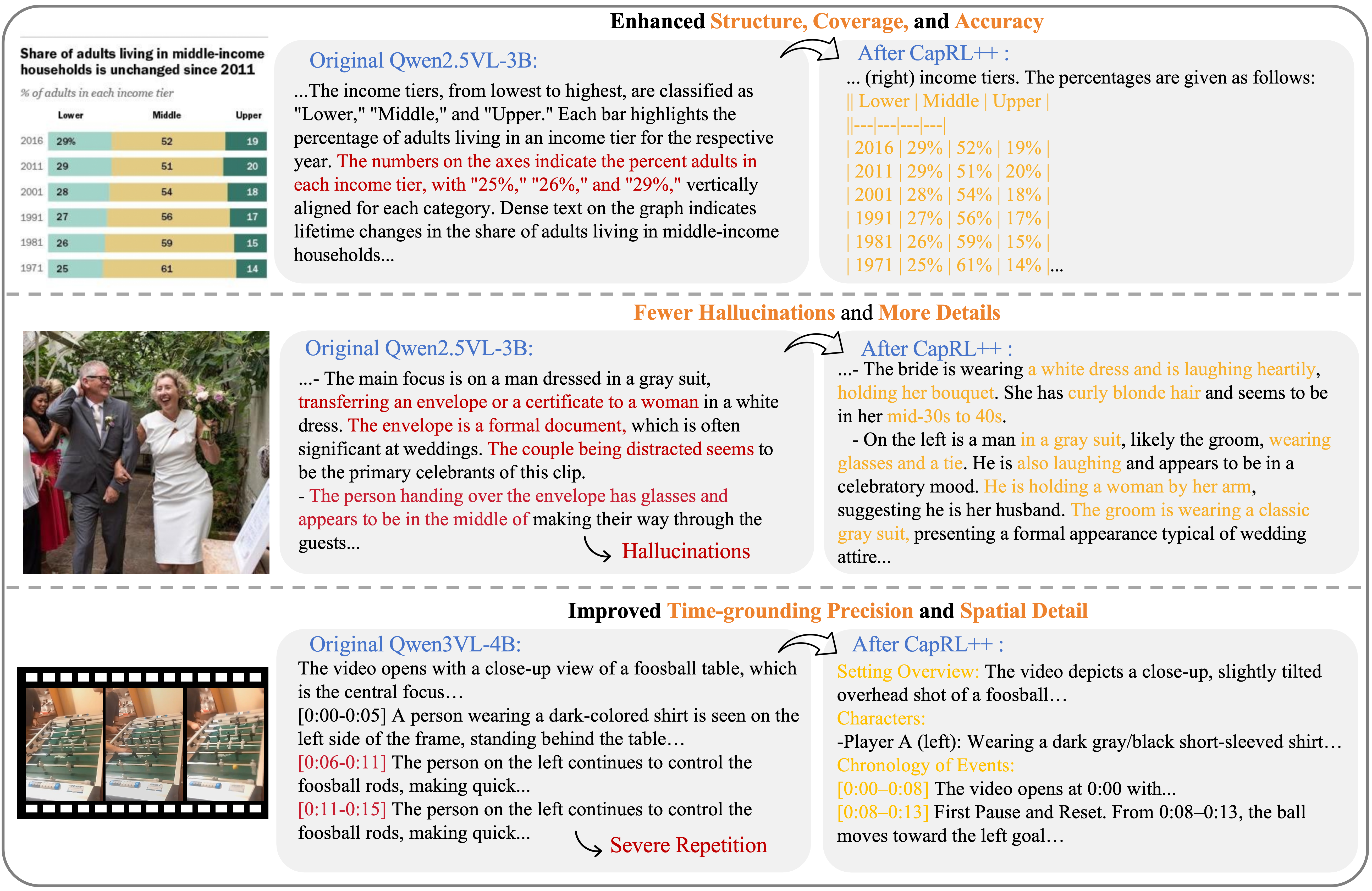}
    \vspace{-1mm}
    \caption{
    \textbf{Qualitative comparison before and after CapRL++ training.} CapRL++ improves caption structure, visual coverage, factual grounding, fine-grained detail, and temporal localization across image and video examples.
    }
    \vspace{-4mm}
    \label{fig:enhance_clarity_case}
\end{figure*}

\section{Related Work}
\label{sec:Related Work}
\noindent \textbf{Image and Video Captioning.}
Early large-scale image–text corpora, such as LAION-400M \cite{schuhmann2022laion}, Conceptual Captions \cite{changpinyo2021conceptual}, and YFCC100M \cite{thomee2016yfcc100m}, have driven significant progress in vision–language pretraining. 
To further improve caption quality, recent works design advanced data curation pipelines. 
For instance, BLIP-LAION \cite{li2022blip} generates synthetic captions, which are subsequently refined or rewritten by large language models as in LaCLIP \cite{fan2023improving}, while approaches such as CapsFusion \cite{yu2024capsfusion} consolidate multi-source information using fine-tuned models. 
Additionally, GPT-4V \cite{openai2023gpt4v_system_card} and human-in-the-loop pipelines have been employed to produce richer and more fine-grained annotations, as demonstrated in ShareGPT4V \cite{chen2024sharegpt4v}, ShareGPT4Video \cite{chen2024sharegpt4video} and ALLaVA \cite{chen2024allava}.
Recent studies \cite{li2024densefusion,sun2024descriptive} further explore multi-expert strategies to compensate for the limitations of individual large vision–language models.

In the video domain, the challenge is further exacerbated by the temporal complexity of videos and the high cost of annotation \cite{abdar2024review,wang2025video,wu2025survey}. 
Large-scale datasets such as YouTube-8M \cite{abu2016youtube}, HowTo100M \cite{miech2019howto100m}, and WebVid-10M \cite{bain2021frozen} rely heavily on weak supervision (e.g., ASR or metadata), resulting in noisy and coarse captions. 
To address the complex multi-modal nature of videos, foundational works explored multi-expert fusion strategies (e.g., InternVideo \cite{wang2022internvideo}) to integrate visual, motion, and audio cues. 
Furthermore, to capture temporal events, dense captioning approaches (e.g., Vid2Seq \cite{yang2023vid2seq}) were introduced to generate multiple segment-level descriptions, though this significantly increases pipeline complexity. 
More recently, to fundamentally improve data quality and reasoning capabilities, advanced methods \cite{maaz2024video,zhang2024llava} extend image-style pipelines to video by leveraging powerful LLMs to rewrite and enrich captions.

In summary, existing approaches in both image and video domains either rely on complex multi-stage pipelines that are training-free but computationally expensive at inference time, 
or require large amounts of high-quality annotated data for supervised fine-tuning. 
In contrast, our CapRL++ achieves strong performance with remarkable data efficiency by leveraging reinforcement learning with verifiable rewards (RLVR), 
without depending on large-scale curated annotations or costly multi-model pipelines.

\noindent \textbf{Reinforcement Learning with Verifiable Rewards (RLVR).} 
RLVR \cite{lambert2024tulu} represents a promising paradigm for training Large Language Models (LLMs) on tasks that have an objective, easily verifiable reward signal.
For example, in mathematical problem-solving, the reward can be a binary signal of correctness \cite{shao2024deepseekmath}, and for code generation, it can be whether the code passes unit tests \cite{team2025kimi}.
Compared with the traditional Supervised Fine-Tuning (SFT), RLVR offers a more robust and scalable approach.
While SFT trains models to imitate a set of provided ground-truth answers, often leading to models that memorize specific phrasings \cite{chu2025sft}, RLVR encourages the model to explore and discover optimal solutions.
This is particularly beneficial for problems with multiple valid answers or reasoning paths.
\section{{Methodology}}
\label{sec:method}

\begin{figure*}[t!]
    \centering     \includegraphics[width=1\textwidth]{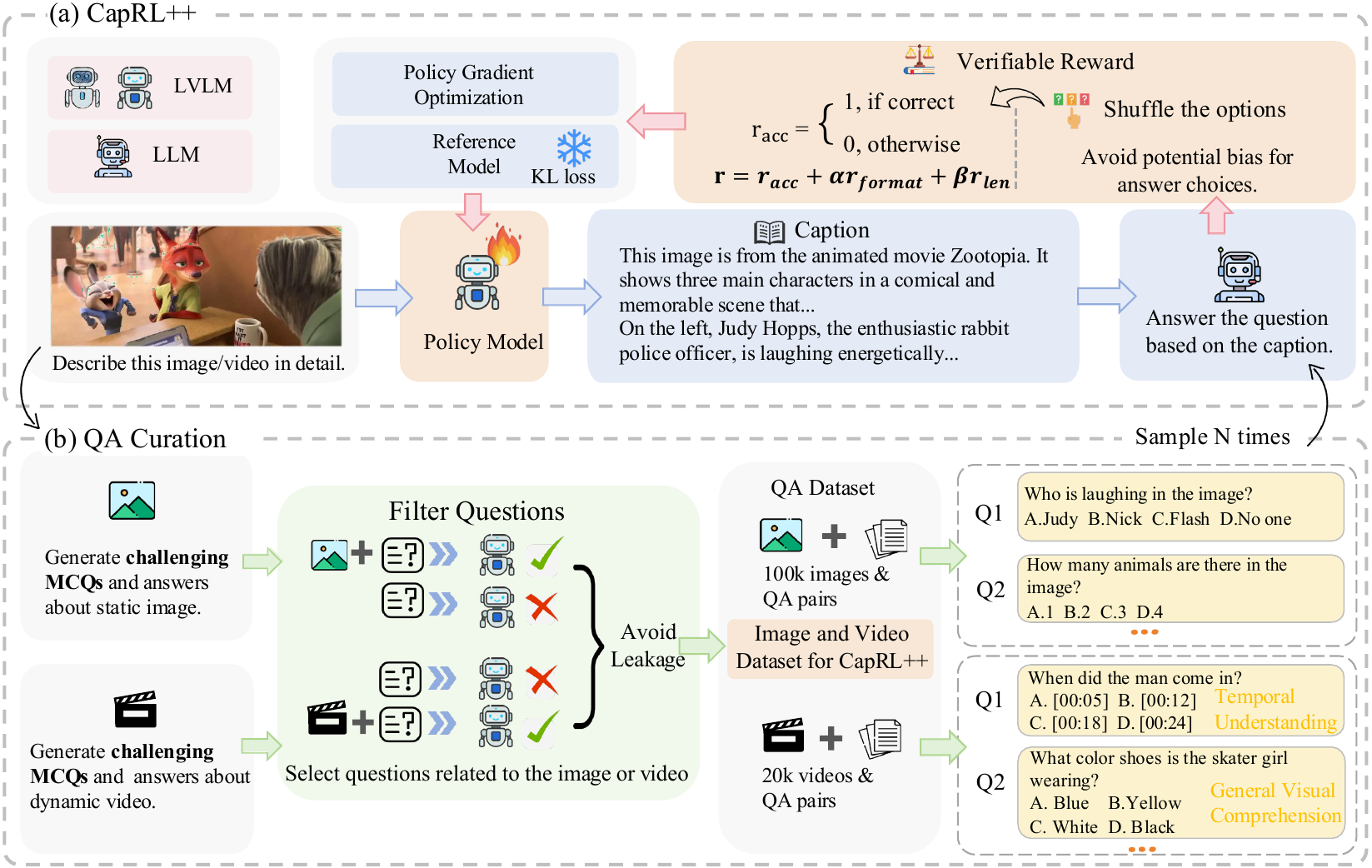}
    \caption{\textbf{Overview of CapRL++.} Unlike the standard single-stage RLVR, our CapRL++ performs a decoupled two-stage process.
    Captions generated by the LVLM, paired with curated MCQs \textbf{(b)}, are used to query an LLM, whose resulting accuracy becomes the objective reward for the LVLM \textbf{(a)}.
    Our CapRL++ offers a scalable framework for applying RLVR to the open-ended image and video captioning task.
    } 
    \vspace{-10pt}
    \label{fig:pipeline}
\end{figure*}


An overview of the \textbf{CapRL++} framework is presented in \cref{fig:pipeline}.
CapRL++ extends RLVR from image captioning to a unified image-video captioning setting, where both static and dynamic visual inputs are optimized through verifiable caption utility.
The framework uses a decoupled two-stage pipeline.
In the first stage, an LVLM generates a caption for an input image or video.
In the second stage, this caption, along with a series of MCQs, is provided as input to a vision-free LLM.
To disentangle spatial perception from temporal reasoning during training, we further introduce \textbf{Spa}tial-Anchored \textbf{Boot}strapping (SpaBoot), which first establishes spatial priors on static images and then expands to dynamic videos.
In the following sections, we will describe how to apply RLVR to the visual captioning task via our CapRL++ in \cref{sec:caprl_unified,sec:reward_design,sec:qa_curation}, and introduce our SpaBoot training recipe in \cref{subsec:curriculum_method}.
\textcolor{black}{Finally, we describe how CapRL++ is used to construct scalable multimodal caption corpora for both images and videos, including the CapRL-Image-5M dataset and the recaptioned CapRL-Video-178K corpus, in \cref{sec:caprl_dataset}.}

\subsection{CapRL++ Overview and Optimization}
\label{sec:caprl_unified}

The design of the reward function is a pivotal factor in the success of RLVR-based approaches, since the reward function directly guides the optimization direction of the policy model.
Although designing reward functions for objective tasks \cite{shao2024deepseekmath,liu2025visual,luo2025gui} is straightforward, developing the reward function for the subjective visual captioning task is challenging. 
While reward models \cite{liu2025inference,su2025crossing,lu2025writing} or the ``LLM-as-a-judge'' approach \cite{gunjal2025rubrics} have been explored for RL training on open-ended tasks, these models are still vulnerable to exploitation in the captioning task, primarily owing to their intrinsic biases, which may unintentionally encourage the captioning model to produce verbose or brief results.
The core philosophy of CapRL++ is that a high-quality visual description should function as a high-fidelity communication carrier.
For both images and videos, we define the quality of a caption $c$ as proportional to its \textit{utility}, the ability to empower a text-only model to perceive what it cannot see.

The overall process of our proposed method CapRL++ is illustrated in \cref{fig:pipeline}.
During the Group Relative Policy Optimization (GRPO) training process, given a visual input $V$ (image or video) and an instruction $X$, the policy model samples a group of candidate captions $\{c_1, c_2, \dots, c_G\}$ .
Each caption is then paired with corresponding questions and fed to a Large Language Model (LLM). 
These captions are validated through a multidimensional reward space to guide the policy optimization toward accuracy, structural adherence, and information efficiency.

To prepare the data for GRPO training, we constructed a VQA dataset composed exclusively of multiple-choice questions.
This multiple-choice format facilitates the computation of verifiable rewards.
Throughout this curation process, we use an LVLM to filter the data and reduce potential language-only leakage.
Further details regarding our reward design and QA curation are provided below.

\subsection{Multidimensional Verifiable Reward Space}
\label{sec:reward_design}

To effectively handle the transition from static 2D scenes to dynamic temporal sequences, we design a suite of verifiable rewards.
The total reward $R_{total}$ for a sampled caption $c_i$ is a weighted combination of three components:

\begin{equation}
R_{total}(c_i) = R_{acc}(c_i) + \alpha R_{format}(c_i) + \beta R_{len}(c_i),
\end{equation}
where $\alpha$ and $\beta$ are hyper-parameters balancing the utility, structural fidelity, and conciseness.

\noindent \textbf{Visual Utility Reward ($R_{acc}$).}
Following the decoupled VQA proxy, each caption $c_i$ is paired with $M$ Multiple-Choice Questions (MCQs) curated for the visual content $V$.
A vision-free LLM $\mathcal{M}_L$ generates answers based \textit{solely} on $c_i$.
To eliminate potential bias in the LLM's preference for specific answer choices, we randomly shuffle the options each time a question is presented.
Additionally, relying on a single answer to evaluate a caption lacks robustness.
To ensure the stability of caption scoring, we sample $N$ times from all the questions related to the image or video and let $\mathcal{M}_L$ answer them independently.
The final reward for a caption is computed as the average accuracy over these $N$ sampled questions:
\begin{equation}
R_{acc}(c_i) = \frac{1}{N} \sum_{k=1}^{N} \mathbb{I} \Big( \mathcal{M}_L(c_i, \operatorname{Shuffle}(q_{m_k})) = \operatorname{GT}_{m_k} \Big),
\end{equation}
where $m_k \sim \{1, \dots, M\}$ and $\operatorname{GT}_{m_k}$ is the ground-truth answer.
This binary correctness signal provides a hard-to-hack optimization target.

\noindent \textbf{Temporal-Anchored Format Reward ($R_{format}$).}
In the video domain, precise temporal localization is essential for high-quality captions. 
Modern base models, such as Qwen3-VL, already demonstrate robust instruction-following 
capabilities for timestamp formatting prior to reinforcement learning. However, since the 
primary objective $R_{acc}$ is heavily focused on information utility, the reinforcement 
learning process could potentially lead to an optimization-induced drift in the output structure. 

To ensure that the model consistently adheres to the desired temporal grounding 
format while pursuing higher utility rewards, we introduce $R_{format}$ as a 
lightweight \textbf{structural constraint}. This term regularizes the model's 
output by evaluating the validity and chronological consistency of the generated timestamps.

Let $\mathcal{S}$ be the sequence of timestamp-like brackets matched via regular expressions 
in the caption $c_i$, and $N_{all} = |\mathcal{S}|$. Let $N_{valid}$ be the number of timestamps 
in $\mathcal{S}$ that conform to logical formatting constraints (e.g., $t_{end} \ge t_{start}$). 
We formulate the format regularization reward as a linear combination:
\begin{equation}
R_{format}(c_i) =  0.5 \cdot v + 0.5\cdot\mathbb{I}_{chrono},
\end{equation}

In this formulation, $v = \frac{N_{valid}}{\max(N_{all}, 1)}$ represents the 
\textbf{validity ratio}, providing a dense signal that rewards the structural integrity of 
the individual time brackets. The indicator function $\mathbb{I}_{chrono}$ evaluates 
\textbf{chronological consistency}, taking the value of 1 if there are valid timestamps 
and their starting times $t_{start}$ are monotonically non-decreasing, and 0 otherwise. 

By incorporating this negligible-cost constraint, we allow the model to stably 
enhance its temporal perception capabilities without compromising its inherent 
ability to produce well-formatted, chronologically sound narrative sequences.

\noindent \textbf{Length-aware Regularization Reward ($R_{len}$).}
To prevent ``reward hacking'' behaviors where the model generates pathologically verbose or redundant text to inflate the probability of covering MCQ answers, we incorporate a length penalty.
Specifically, we adopt a simple yet effective truncation-based penalty or a linear cost function \cite{chen2025avocado}:

\begin{equation}
R_{len}(c_i) =
\begin{cases}
1, & \text{if } \operatorname{len}(c_i) \leq \tau_{\text{1}}, \\
1-\frac{\operatorname{len}(c_i) - \tau_{\text{1}}}{\tau_{\text{2}} - \tau_{\text{1}}}, & \text{if } \tau_{\text{1}} < \operatorname{len}(c_i) \leq \tau_{\text{2}}, \\
0, & \text{if } \operatorname{len}(c_i) > \tau_{\text{2}}.
\end{cases}
\end{equation}
where $\tau_{\text{1}}$ and $\tau_{\text{2}}$ are predefined token budgets.
This encourages the model to express useful information concisely, thereby improving information efficiency.
In our experiments, we set $\tau_{\text{1}} = 2048$ and $\tau_{\text{2}} = 3072$.

\subsection{Multimodal MCQ Curation for Reward Construction}
\label{sec:qa_curation}

Reliable reward signals depend on high-quality MCQ data.
We construct this VQA dataset using a structured three-stage curation pipeline as shown in \cref{fig:pipeline}.

\noindent\textbf{(1) Image and Video Collection.}
For images, we source diverse samples from the web and existing open-source datasets, spanning natural scenes, charts, and documents to maximize visual variety.
For videos, we select short clips (mostly under 30\,s) from the publicly available dataset \textbf{LLaVA-Video-178K} \cite{zhang2024llava} for training.
The model's generalization capability to long videos is further evaluated during inference.

\noindent\textbf{(2) QA Generation.}
We employ a mixture of strong LVLMs, including Qwen3-VL-235B-A22B and Gemini-3-Flash, to automatically generate MCQs for each image and video.
Leveraging multiple generator models effectively mitigates potential annotation biases that would arise from relying on a single model.
Notably, the construction of video QA data poses unique challenges compared to its image counterpart, as it must capture not only static visual attributes but also dynamic, temporally evolving content.
To address this, we carefully design two complementary categories of video MCQs:
\textbf{(i) Temporal localization questions}, which probe the model's ability to identify precise event boundaries, temporal ordering, and interval durations using explicit timestamp references (e.g., \textit{``At what timestamp does the man first open the blue door?''});
and \textbf{(ii) General visual comprehension questions}, which target holistic understanding of subjects, actions, scenes, object relationships, and fine-grained visual details (e.g., \textit{``What object is the woman holding in her left hand?''}).
This dual-category design ensures that the resulting reward signals encourage the model to produce captions that are both temporally precise and semantically comprehensive.

\noindent\textbf{(3) QA Filtering.}
Finally, we implement a stringent filtering process to guarantee that all questions are strictly visually grounded and answerable exclusively through the analysis of the visual content.
This step is crucial to prevent information leakage and ensures that the model must perform genuine visual understanding, rather than exploiting external knowledge or superficial cues embedded within the question itself \cite{ma2024robust}.
Specifically, the filtered set of QA pairs, denoted as $\mathcal{Q}$, is defined as:
\begin{equation}\label{eq:filtering}
    \mathcal{Q} = \{(q, a) \in \mathcal{D} \mid \mathcal{M}_{V_f}(q, I) = a \;\land\; \mathcal{M}_{V_f}(q) \ne a\},
\end{equation}
where $(q, a)$ is a question-answer pair from the initial generated dataset $\mathcal{D}$,
$I$ is the corresponding input image or video,
$\mathcal{M}_{V_f}$ is the LVLM used for QA filtering,
$\mathcal{M}_{V_f}(q, I)$ denotes the answer produced when conditioned on both the question $q$ and the visual input $I$,
and $\mathcal{M}_{V_f}(q)$ denotes the answer produced when the visual input is omitted.
This bidirectional verification ensures that every retained question genuinely requires visual perception, thereby providing reliable and unbiased reward signals for RL training.

In the preliminary version of this work, we constructed a foundational dataset of approximately \textbf{75K} image-QA pairs to train the initial image-caption model based on Qwen2.5-VL-3B.
To support the extension into the video domain and to fully leverage the capabilities of the more advanced Qwen3-VL backbone, we comprehensively expanded and reconstructed our training corpus.
The updated image dataset consists of \textbf{100K} high-quality image-QA pairs sourced from diverse domains, while the video dataset comprises about \textbf{20K} carefully curated video-QA pairs filtered from LLaVA-Video-178K \cite{zhang2024llava}.
The inclusion of high-quality video QA data is particularly critical: the temporal localization questions enforce accurate timestamp grounding, while the general comprehension questions encourage holistic scene understanding, together driving the model to generate captions that are temporally faithful, semantically rich, and firmly grounded in the visual evidence.

\begin{figure}[t!]
  \centering
  \includegraphics[width=0.8\columnwidth]{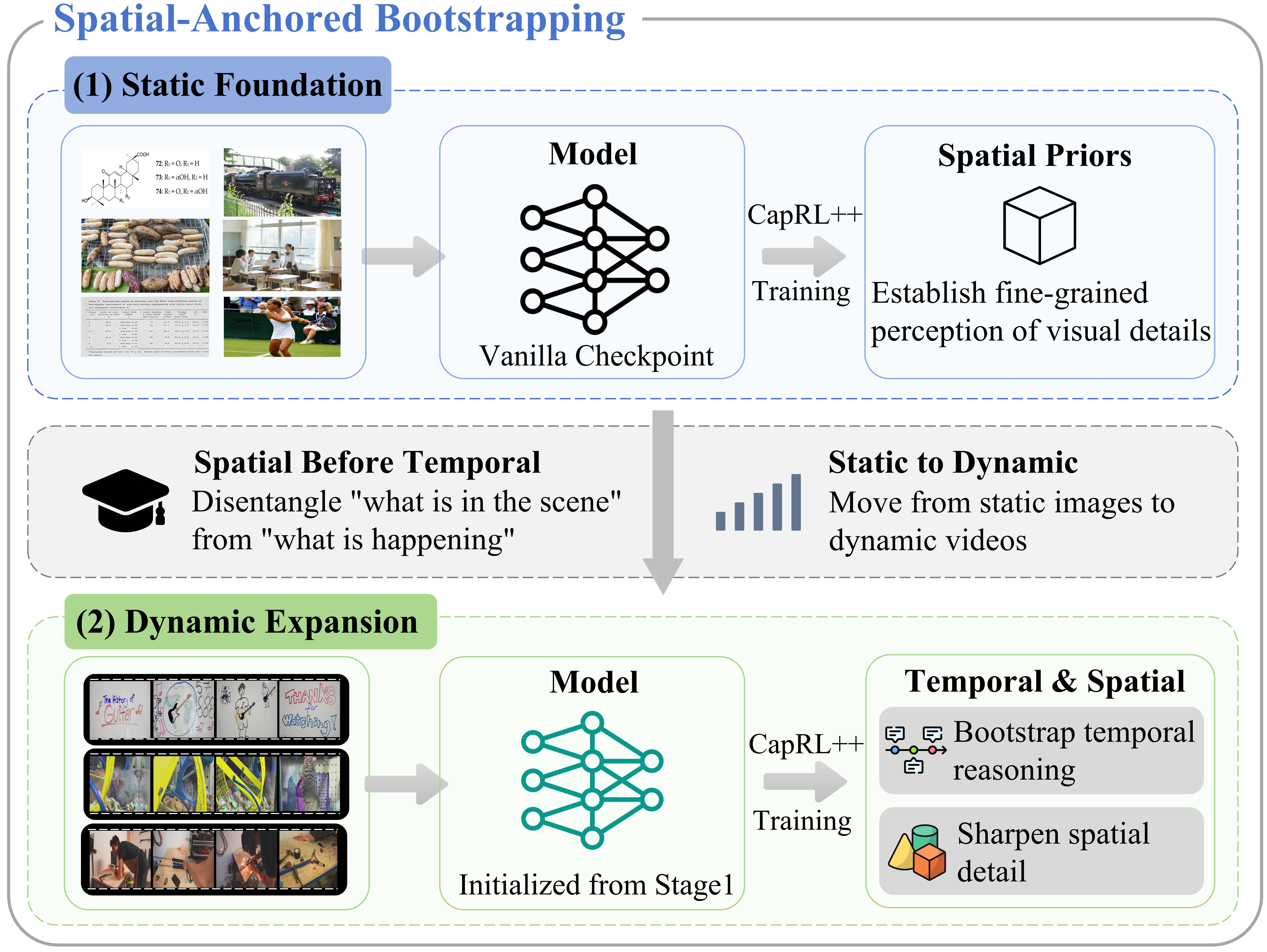}
  \vspace{-5pt}
  \caption{Overview of the Spatial-Anchored Bootstrapping. Stage~1 establishes spatial priors on static images; Stage~2 initializes from Stage~1 and trains on videos to bootstrap temporal reasoning while further sharpening spatial detail.
  }
  \vspace{-15pt}
  \label{fig:curriculum_pipeline}
\end{figure}

\subsection{Spatial-Anchored Bootstrapping}
\label{subsec:curriculum_method}
Video captioning is intrinsically a coupled problem: a single model must identify \emph{what is in the scene} and reason about \emph{what is happening} under one reward signal. Training directly on dynamic content forces these two capabilities to emerge jointly, which in practice makes the reward signal ambiguous. To disentangle these two factors, we propose \textbf{Spa}tial-Anchored \textbf{Boot}strapping (SpaBoot), a two-stage recipe that establishes a spatial perception foundation before introducing temporal understanding.

\begin{figure*}[t]
  \centering
  \includegraphics[width=0.99\textwidth]{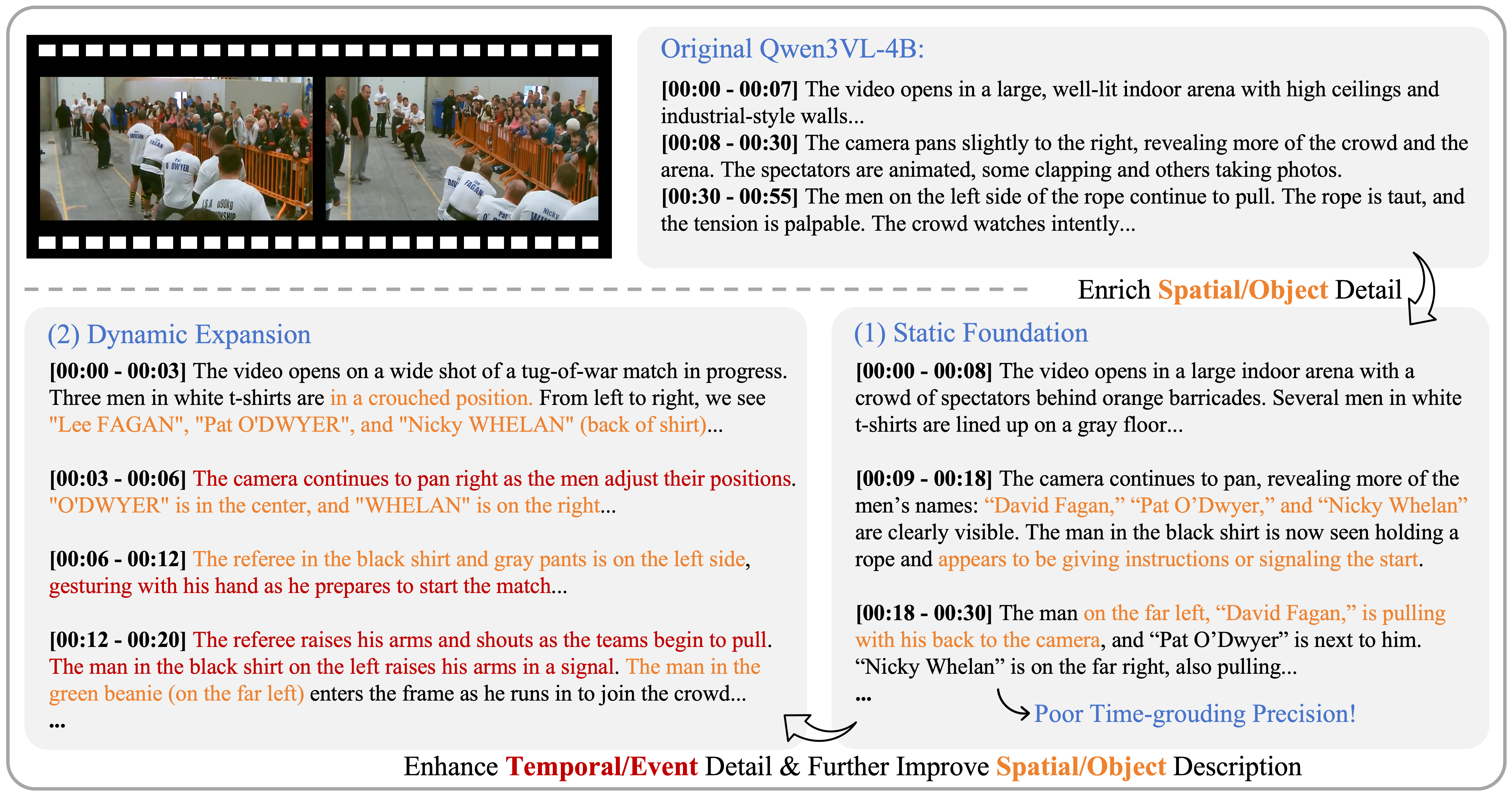}
  \vspace{-0.5mm}
  \caption{
  \textbf{Qualitative analysis of progressive captioning improvements through SpaBoot training.}
  }
  \vspace{-4mm}
  \label{fig:case_study}
\end{figure*}

\noindent \textbf{(1) Stage 1: Static Foundation.} The model is first trained exclusively on static images using the utility reward ($R_{acc}$) and length penalty ($R_{len}$). This stage forces the model to focus on developing strong spatial priors, mastering capabilities such as fine-grained attribute extraction and precise Optical Character Recognition (OCR). \textbf{(2) Stage 2: Dynamic Expansion.} Initialized from the Stage-1 checkpoint, the model is subsequently trained on video data. The full multidimensional reward space is activated, including the temporal-anchored format reward ($R_{format}$). Equipped with a robust spatial foundation, the model can efficiently allocate its optimization capacity toward learning temporal dynamics. At the same time, exposure to richer visual contexts further sharpens spatial detail.

This progressive paradigm allows the model to smoothly layer complex temporal comprehension (e.g., precise temporal segmentation and event ordering) on top of highly accurate spatial representations. 

\subsection{CapRL++-Annotated Image and Video Corpora}

\label{sec:caprl_dataset}
We use the resulting CapRL++ captioner to curate large-scale, high-quality multimodal pretraining datasets. 
Existing data curation pipelines often rely on costly proprietary models or complex multi-expert filtering. 
In contrast, our CapRL++ paradigm enables a relatively lightweight model to generate highly reliable, dense descriptions across both spatial and temporal domains at a fraction of the cost.

\noindent \textbf{CapRL-Image-5M Dataset.}
For the static visual domain, we construct a 5M image-caption dataset. 
We combine large-scale open-source datasets (ShareGPT4V-1M and DenseFusion-1M) with 3M web-crawled images. 
Given the inherent noise and safety risks of web images, we apply a rigorous three-stage filtering pipeline:
(1) \textbf{Quality and safety screening}, where we remove semantically redundant, low-resolution, and unsafe content (e.g., violence or pornography) using clustering techniques inspired by SemDeDup \cite{abbas2023semdedup}; 
(2) \textbf{Benchmark decontamination}, which eliminates images exhibiting high similarity to standard evaluation sets to rigorously prevent data leakage; and 
(3) \textbf{Human verification}, involving sample-based screening to ensure strict safety compliance.
This process yields 3M high-quality web images.
All 5M images are then densely annotated using Qwen2.5-VL-3B trained with CapRL++. 

\noindent \textbf{CapRL-Video-178K Dataset.}
For the dynamic visual domain, we enhance the LLaVA-Video-178K \cite{zhang2024llava} dataset.
As this corpus has already undergone rigorous quality filtering in its original construction, we directly apply Qwen3-VL-4B trained with CapRL++ to generate dense, high-fidelity captions.
This process replaces the original GPT-4o \cite{hurst2024gpt} annotations with utility-driven descriptions that better capture fine-grained temporal dynamics.
\section{Experiments}
\label{sec:exp}

To comprehensively assess the caption quality of CapRL++, we organize empirical study along three axes.
We first evaluate CapRL++ on the \textbf{image domain} (\cref{subsec:image_experiments}), where captions are examined both as pretraining data for LVLMs and as standalone descriptions on caption-quality benchmarks.
We then extend the evaluation to the \textbf{video domain} (\cref{subsec:video_experiments}), which additionally probes temporal perception and long-horizon understanding.
Finally, we conduct extensive \textbf{discussion and ablation studies} (\cref{subsec:discussion_and_ablations}) to analyze cross-modal transferability, scaling behavior, and key design choices.


\subsection{Experimental Setup}
\label{subsec:experimental_setup}

\noindent \textbf{Caption Evaluation Strategy.}
We assess caption quality from two complementary perspectives that apply uniformly to image and video captions.
Given that the dominant downstream use of caption data in LVLMs is continual pretraining (CPT), we first measure the \emph{downstream utility} of our caption dataset under the CPT setting.
Second, we evaluate the \emph{intrinsic quality} of the captions themselves, focusing on informativeness and faithfulness through the Prism framework \cite{qiao2024prism} together with other dedicated captioning benchmarks (e.g., HAT \cite{petryk2024aloha} for images and Dream-1K \cite{wang2024tarsier} and CaReBench \cite{xu2024carebench} for videos).
 


\noindent \textbf{Backbone Selection.}
For image experiments, we build upon the conference version of this work and adopt the Qwen2.5-VL architecture as our captioner, which demonstrates strong and consistent improvements across all settings.
For video experiments, we transition to the more capable Qwen3-VL backbone to better handle the additional complexity of temporal reasoning.
This architectural progression naturally reflects the increasing demands when extending CapRL++ from static images to dynamic videos.


\noindent \textbf{Evaluation Benchmarks.}
Our evaluation covers a broad range of benchmarks for both images and videos.
For image benchmarks, we consider document and chart understanding benchmarks
(e.g., InfoVQA \cite{mathew2022infographicvqa}, DocVQA \cite{mathew2021docvqa}, ChartQA \cite{masry2022chartqa}, ChartQAPro \cite{masry2025chartqapro} and CharXiv \cite{wang2024charxiv}); 
scientific, mathematical, and diagram reasoning benchmarks, 
(e.g., MathVista \cite{lu2024mathvista}, MathVerse \cite{zhang2024mathverse}, MathVision \cite{wang2024measuring}, WeMath \cite{qiao2025we}, MM-Vet \cite{yu2023mm}, and AI2D \cite{kembhavi2016diagram});
and general multimodal perception and reasoning benchmarks,
(e.g., RealWorldQA \cite{xai_org_realworldqa}, SEED2Plus \cite{li2024seed}, MME-RealWorld \cite{yu2023mm}, MMBench \cite{liu2024mmbench}, MMStar \cite{chen2024we}, GQA \cite{hudson2019gqa}, SEED-Bench \cite{li2023seed}, MMMU \cite{yue2024mmmu}, and MMMU-Pro \cite{yue2025mmmu}).
We also evaluate on the traditional caption-quality benchmark HAT (HAllucination Test) \cite{petryk2024aloha}
using specific metrics.

For video benchmarks, we consider general video understanding benchmarks, 
including Video-MME \cite{fu2025video}, MMVU \cite{zhao2025mmvu}, MVBench \cite{li2024mvbench}, and Tomato \cite{shangguan2025tomato}; 
temporal and spatial reasoning benchmarks, including TempCompass \cite{liu2024tempcompass}, MotionBench \cite{hong2025motionbench}, and TimeLens-Bench \cite{zhang2025timelens},
which specifically targets temporal grounding with instances re-annotated from Charades-STA \cite{gao2017tall}, QVHighlights \cite{lei2021detecting}, and ActivityNet Captions \cite{krishna2017dense}; 
long-video understanding benchmark LongVideoBench \cite{wu2024longvideobench}; 
and caption-quality benchmarks, including Dream-1K \cite{wang2024tarsier} and CaReBench \cite{xu2024carebench}.
\subsection{Image Experiments}
\label{subsec:image_experiments}

\subsubsection{Evaluation via Continual Pretraining}
\label{subsubsec:image_pretraining}

To thoroughly evaluate the quality of captions produced by CapRL++ as pretraining data, we conduct comprehensive comparisons with widely used caption datasets from the open-source community.

\begin{table*}[!t]
  \centering
  \scriptsize
  \caption{\textbf{Performance comparison using different pretraining datasets.}
  CapRL-Image-1M consistently outperforms other datasets across all three model configurations, and further improvements are observed when scaling to 5M.
  The best results are \textbf{bold} and the second-best are \underline{underlined}.
  }
  \vspace{0pt}
  \label{tab:Main Table}
  \setlength{\tabcolsep}{3.5pt}
  \begin{tabular*}{\textwidth}{@{\extracolsep{\fill}}l@{\hspace{3pt}\vrule\hspace{3pt}}ccccccccccccc}
    \toprule
    \makecell{Pretraining\\Dataset} & \makecell{Info\\VQA} & \makecell{Doc\\VQA} & \makecell{Chart\\QA} & \makecell{Real\\WorldQA} & \makecell{Math\\Vista} & \makecell{SEED2\\Plus} & \makecell{MME\\RW} & \makecell{MMB} & MMStar & MMVet & AI2D & GQA & Average \\
    \midrule

    \rowcolor{gray!10}
    \multicolumn{14}{c}{\textit{Qwen2.5-3B + Qwen2.5-ViT}}\\
    Vanilla           & 43.9 & 81.0 & 72.7 & 55.1 & 41.6 & 56.6 & 30.5 & 68.6 & 44.7 & 41.0 & 68.3 & 61.5 & 55.5 \\
    ShareGPT4V-1M     & 46.1 & 82.4 & 74.2 & 55.0 & 44.7 & 60.5 & 29.8 & 68.9 & 45.2 & 42.4 & 70.1 & 61.4 & 56.7 \\
    DenseFusion-1M    & 49.4 & 84.6 & 74.4 & 54.1 & 44.6 & 59.1 & \underline{30.7} & 69.0 & 45.6 & 40.2 & 70.4 & \underline{62.5} & 57.1 \\
    CapRL-Image-1M          & \underline{56.2} & \underline{87.3} & \underline{78.0} & \underline{55.1} & \underline{45.5} & \underline{62.0} & 30.3 & \underline{70.5} & \underline{47.0} & \underline{50.0} & \underline{72.9} & 61.6 & \underline{59.7} \\
    CapRL-Image-5M          & \textbf{61.5} & \textbf{90.0} & \textbf{80.5} & \textbf{57.6} & \textbf{48.1} & \textbf{63.2} & \textbf{30.9} & \textbf{73.1} & \textbf{50.4} & \textbf{52.6} & \textbf{74.7} & \textbf{62.6} & \textbf{62.0} \\
    \midrule

    \rowcolor{gray!10}
    \multicolumn{14}{c}{\textit{Qwen2.5-7B + Qwen2.5-ViT}}\\
    Vanilla           & 47.6 & 83.7 & 77.1 & 55.9 & 47.4 & 60.4 & 29.4 & 72.1 & 48.1 & 47.1 & 72.4 & 62.7 & 58.7 \\
    ShareGPT4V-1M     & 49.8 & 85.1 & 75.7 & 56.8 & 46.6 & 60.9 & 31.8 & 71.9 & 48.4 & 45.9 & 72.2 & 62.7 & 59.0 \\
    DenseFusion-1M    & 53.5 & 87.8 & 76.7 & 58.6 & 46.3 & 61.0 & 31.1 & \underline{72.6} & 48.6 & 49.7 & 72.5 & 63.1 & 60.2 \\
    CapRL-Image-1M          & \underline{59.9} & \underline{89.5} & \underline{80.6} & \underline{58.9} & \underline{50.4} & \underline{63.1} & \underline{32.2} & 72.1 & \underline{51.3} & \underline{50.5} & \underline{75.3} & \underline{63.2} & \underline{62.2} \\
    CapRL-Image-5M          & \textbf{63.4} & \textbf{91.4} & \textbf{81.5} & \textbf{61.4} & \textbf{50.8} & \textbf{63.2} & \textbf{34.9} & \textbf{72.7} & \textbf{52.6} & \textbf{52.6} & \textbf{76.9} & \textbf{63.8} & \textbf{63.8} \\
    \midrule

    \rowcolor{gray!10}
    \multicolumn{14}{c}{\textit{InternLM2.5-7B + CLIP-ViT-L}}\\
    Vanilla           & 37.4 & 73.2 & 68.7 & 56.9 & 44.2 & 58.2 & 30.7 & 70.7 & 47.0 & 43.1 & 71.8 & 64.9 & 55.6 \\
    ShareGPT4V-1M     & 38.9 & 73.8 & 69.8 & 56.3 & 44.8 & 59.9 & 33.2 & 72.6 & 46.2 & 43.3 & 72.7 & 65.0 & 56.4 \\
    DenseFusion-1M    & 39.3 & 76.4 & 70.8 & \textbf{59.7} & 44.5 & 60.3 & 34.1 & 72.2 & 47.9 & 44.0 & 73.7 & 65.5 & 57.4 \\
    CapRL-Image-1M          & \underline{43.3} & \underline{80.0} & \underline{75.8} & \underline{58.0} & \underline{49.6} & \underline{62.8} & \underline{34.1} & \underline{73.4} & \underline{50.2} & \underline{46.6} & \underline{76.0} & \underline{65.8} & \underline{59.6} \\
    CapRL-Image-5M          & \textbf{47.0} & \textbf{83.5} & \textbf{77.7} & \textbf{59.7} & \textbf{50.4} & \textbf{63.5} & \textbf{38.9} & \textbf{73.7} & \textbf{53.3} & \textbf{54.3} & \textbf{77.6} & \textbf{66.3} & \textbf{62.2} \\
    \bottomrule
  \end{tabular*}
  \vspace{-2mm}
\end{table*}

\noindent \textbf{Implementation details.}
In our setup, the language model is initialized with a pretrained LLM, the visual encoder with a pretrained ViT, and the MLP projector is randomly initialized, following a standard multimodal pretraining scheme.
We conduct experiments under three settings: Qwen2.5-3B + Qwen2.5-ViT, Qwen2.5-7B + Qwen2.5-ViT, and InternLM2.5-7B + CLIP-ViT-L.
Training follows the ShareGPT4V paradigm in three stages: Initial Alignment with BLIP-558K \cite{li2022blip} dataset; Further Pretraining with diverse high-quality image-caption datasets; and SFT with Open-LLaVA-NeXT-1M \cite{chen2024open}.
For comparison, we adopt strong baselines including Vanilla, which skips Further Pretraining, ShareGPT4V-1M, DenseFusion-1M, and CapRL-Image-1M (randomly sampled from CapRL-Image-5M).

\noindent \textbf{Main results.}
As shown in \cref{tab:Main Table}, CapRL-Image-1M consistently outperforms both ShareGPT4V-1M and DenseFusion-1M across most benchmarks under all three model configurations.
Under the Qwen2.5-3B + Qwen2.5-ViT setting, it improves over DenseFusion-1M by 6.8\%, 2.7\%, and 3.6\% on InfoVQA, DocVQA, and ChartQA, respectively, suggesting that CapRL++ captions are particularly beneficial for fine-grained and structured visual understanding.
CapRL-Image-1M also brings consistent gains on general perception benchmarks such as MMStar and MMBench, indicating that high-quality dense captions improve both domain-specific and natural-image understanding.
Scaling from CapRL-Image-1M to CapRL-Image-5M further improves the average performance across all settings, demonstrating the scalability and practical value of CapRL++ for constructing high-quality multimodal pretraining data.

\begin{table*}[!t]
  \centering
  \scriptsize
  \caption{\textbf{Ablation on image sources.}
  We annotate the images in ShareGPT4V-1M and DenseFusion-1M using Qwen2.5-VL-3B trained with CapRL++, and use them respectively as pretraining datasets for comparison.}
  \vspace{0pt}
  \label{tab:ablation_image_sources}
  \setlength{\tabcolsep}{2.2pt}
  \begin{tabular*}{\textwidth}{@{\extracolsep{\fill}}l@{\hspace{3pt}\vrule\hspace{3pt}}ccccccccccccc}
    \toprule
    \makecell{Pretraining\\Dataset} & \makecell{Info\\VQA} & \makecell{Doc\\VQA} & \makecell{Chart\\QA} & \makecell{Real\\WorldQA} & \makecell{Math\\Vista} & \makecell{SEED2\\Plus} & \makecell{MME\\RW} & \makecell{MMB} & MMStar & MMVet & AI2D & GQA & Average \\
    \midrule

    \rowcolor{gray!10}
    \multicolumn{14}{c}{\textit{Qwen2.5-3B + Qwen2.5-ViT}}\\
    Vanilla             & 43.9 & 81.0 & 72.7 & 55.1 & 41.6 & 56.6 & 30.5 & 68.6 & 44.7 & 41.0 & 68.3 & 61.5 & 55.5 \\
    \midrule

    ShareGPT4V-1M       & 46.1 & 82.4 & 74.2 & 55.0 & 44.7 & \textbf{60.5} & 29.8 & 68.9 & 45.2 & 42.4 & 70.1 & 61.4 & 56.7 \\
    CapRL-ShareGPT4V-1M & \textbf{52.1} & \textbf{85.9} & \textbf{75.2} & \textbf{56.3} & \textbf{45.6} & 60.0 & \textbf{30.9} & \textbf{70.9} & \textbf{46.7} & \textbf{47.5} & \textbf{71.4} & \textbf{61.7} & \textbf{58.7} \\
    \midrule

    DenseFusion-1M      & 49.4 & 84.6 & 74.4 & 54.1 & 44.6 & 59.1 & 30.7 & 69.0 & 45.6 & 40.2 & 70.4 & \textbf{62.5} & 57.1 \\
    CapRL-DenseFusion-1M & \textbf{55.0} & \textbf{87.8} & \textbf{77.5} & \textbf{56.2} & \textbf{44.7} & \textbf{62.8} & \textbf{32.0} & \textbf{71.0} & \textbf{46.6} & \textbf{49.9} & \textbf{72.7} & 62.3 & \textbf{59.9} \\
    \bottomrule
  \end{tabular*}
  \vspace{-5pt}
\end{table*}

\noindent \textbf{Ablations about image sources.} 
In the previous comparisons, the images used in each dataset are not identical. 
To better control for this variable, we fix the set of images and instead compare the effect of caption quality of different datasets under the Qwen2.5-3B + Qwen2.5-ViT setting. 
As shown in \cref{tab:ablation_image_sources}, we compare CapRL++ with ShareGPT4V-1M and DenseFusion-1M. 
The results demonstrate that, when using the same set of images, further pretraining with the CapRL++-annotated dataset enables the LVLM to outperform the baselines by more than 2\%. 
This finding indicates that the substantial advantage of the CapRL++ dataset over the baselines largely stems from the superior quality of its captions, rather than from differences in image diversity.

\begin{figure*}[!t]
  \centering
  \includegraphics[width=0.99\textwidth]{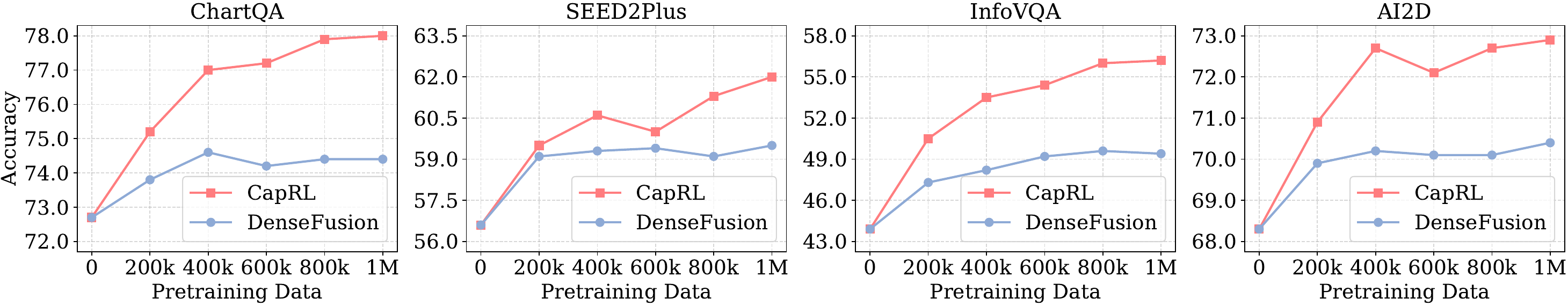}
  \vspace{-1mm}
  \caption{
  The scaling performance comparison between CapRL-Image-1M and DenseFusion-1M. We use different amounts of pretraining data from the two datasets to observe the scaling trend. 
  }
  \vspace{-4mm}
  \label{imgs:caprl_densefusion_scaling}
\end{figure*}

\noindent \textbf{Scaling trend comparison of different datasets.} 
We further compare the scaling trend of CapRL++ and DenseFusion under the Qwen2.5-3B + Qwen2.5-ViT setting. 
Specifically, we sample different numbers of image-caption pairs from each dataset for pretraining. 
As shown in \cref{imgs:caprl_densefusion_scaling}, the CapRL++ dataset consistently outperforms the corresponding DenseFusion dataset across various scales of pretraining data. 
Moreover, the overall trend indicates that this performance gap continues to widen as the data size increases.
This phenomenon highlights the strong scaling properties of the CapRL++ dataset: thanks to its high-quality captions, LVLMs continue to benefit as the dataset size grows. 

\subsubsection{Evaluation via Image Caption Quality Benchmarks}

\label{subsubsec:image_caption_evaluation}

In the previous section, we demonstrated from the pretraining perspective that image captions generated by CapRL++ are highly beneficial for modality alignment.
In this section, we directly evaluate the informativeness of the captions produced by CapRL++ through downstream benchmarks, and compare them against other captioning models.


\noindent \textbf{Evaluation via the Prism framework.}
We evaluate captioning performance through the lens of the Decoupled VQA in the Prism Framework \cite{qiao2024prism}, which evaluates caption quality in an objective and stable manner. Conceptually aligned with our QA-driven reward design, Prism decouples the VQA process into two distinct stages. In Stage 1, the evaluated LVLM acts as a captioner to generate a detailed description of the input image or video. In Stage 2, a text-only LLM answers task-specific questions based solely on the generated caption as context. In our experimental setup, we fix the Stage 2 answering LLM to a fine-tuned Qwen2.5-3B-Instruct. This strict control variable ensures that the final benchmark performance directly and exclusively reflects the descriptive quality and perceptual grounding of the captions produced by the Stage 1 model.

\begin{table*}[!t]
  \centering
  \scriptsize
  \caption{\textbf{Image captioning ability comparison in the Prism Framework. }CapRL++ (based on Qwen2.5-VL-3B) achieves comparable performance to Qwen2.5-VL-72B, and significantly surpasses existing strategies that use LVLM-as-a-Judge as the reward. The best results are \textbf{bold} and the second-best results are \underline{underlined}.}
  \vspace{-1mm}
  \label{tab:Prism main table}
  \resizebox{\textwidth}{!}{%
  \setlength{\tabcolsep}{2.5pt}
  \begin{tabular}{l|c|cccccccccccc}
    \toprule
    \makecell{Caption\\Model} & \makecell{GRPO\\Trained} & \makecell{Chart\\QA} & \makecell{ChartQA\\Pro} & \makecell{Info\\VQA} & \makecell{MMMU\\Pro} & \makecell{Math\\Verse} & \makecell{Char\\Xiv} & \makecell{We\\Math} & \makecell{Math\\Vision} & MMStar & SEED & MMMU & Average \\
    \midrule
    Qwen2.5-VL-3B (\textit{baseline}) & \ding{55} & 65.6 & 27.1 & 40.2 & 28.6 & 32.8 & 21.8 & 54.4 & 22.6 & 46.4 & 64.1 & 35.1 & 39.9 \\
    Qwen2.5-VL-7B  & \ding{55} & 74.9 & 35.4 & 56.4 & 30.1 & 36.4 & 24.8 & 57.0 & 23.3 & 50.7 & 67.1 & 37.9 & 44.9 \\
    Qwen2.5-VL-72B & \ding{55} & \underline{80.2} & \underline{38.0} & \underline{60.8} & \textbf{34.1} & \textbf{39.9} & \underline{30.7} & \textbf{60.2} & \textbf{24.5} & \textbf{55.0} & \underline{69.3} & \textbf{39.4} & \textbf{48.3} \\
    UnifiedRW-as-Judge-3B    & \ding{51} & 54.9 & 25.1 & 33.6 & 28.1 & 34.6 & 20.4 & 58.2 & 24.5 & 45.4 & 61.2 & 36.3 & 38.4 \\
    Qwen2.5-VL-as-Judge-3B    & \ding{51} & 71.4  & 34.2   & 49.3   & 29.1   & 33.8   & 22.9   & 54.3   & 24.1   & 47.7   & 64.5   & 36.4   & 42.5   \\
    \rowcolor[HTML]{DAEFF9}
    Ours (w/ CapRL++)        & \ding{51} & \textbf{80.5} & \textbf{39.9} & \textbf{64.8} & \underline{30.7} & \underline{36.4} & \textbf{32.4} & \underline{60.1} & \underline{23.4} & \textbf{55.0} & \textbf{70.6} & \underline{38.1} & \textbf{48.3} \\
    \bottomrule
  \end{tabular}
  }%
\vspace{-1mm}
\end{table*}

As shown in \cref{tab:Prism main table}, CapRL++ yields substantial gains on the Qwen2.5-VL series.
Starting from Qwen2.5-VL-3B, the resulting model significantly outperforms both the 3B and 7B base models, and achieves performance comparable to Qwen2.5-VL-72B.
In particular, it improves over the 3B baseline by 14.9\%, 12.8\%, and 24.6\% on ChartQA, ChartQAPro, and InfoVQA, respectively, and by 8.6\% and 6.5\% on MMStar and SEED.
These results suggest that GRPO training with CapRL++ can effectively unlock the visual understanding potential of Qwen2.5-VL-3B by encouraging the model to organize objects, attributes, and relations in the image into more comprehensive captions, thereby improving downstream perception performance.

\noindent \textbf{Comparison with LVLM-as-a-Judge reward.}
We further compare CapRL++ with alternative reward designs based on LVLM-as-a-Judge on the Qwen2.5-VL-3B backbone.
When using UnifiedReward-2.0-qwen-3b \cite{wang2025unified} as the judge to evaluate caption quality, the model's captioning ability actually deteriorates during GRPO training.
We attribute this to a strong bias in UnifiedReward-2.0-qwen-3b: during training, it was exposed to many captions from text-to-image datasets, which are typically short and focus only on the main objects.
As a result, the reward model tends to favor overly short captions.
As shown in \cref{fig:teaser}, the average caption length during training continuously decreases and eventually collapses to producing only ``:description''.
In contrast, when using Qwen2.5-VL-3B itself as the judge, the bias goes in the opposite direction: it tends to favor overly verbose captions.
This makes the policy model prone to exploiting the reward by generating long passages that are only weakly grounded in the image, yet still satisfy the judge model's preference.
As shown in \cref{tab:Prism main table}, both LVLM-as-a-Judge variants perform substantially worse than CapRL++.
Overall, these observations suggest that LVLM-as-a-Judge rewards can be unreliable in practice due to strong and hard-to-control annotation biases.
In contrast, CapRL++ converts subjective caption preference into more objective, task-oriented assessments, leading to more stable and effective optimization.

\noindent \textbf{Evaluation on more captioning benchmarks.}
Beyond the Prism framework, we also evaluate CapRL++ on other benchmarks tailored for caption evaluation.
We conduct extensive evaluations on HAT using ALOHa \cite{petryk2024aloha}, CLAIR \cite{chan2023clair}, and the Factuality and Coverage metrics proposed by CapMAS \cite{lee2024toward}. 
These four metrics rely on GPT-based judging or specialized match algorithms to provide automated, quantitative assessments of caption quality.
As shown in \cref{tab:hat_case}, CapRL++ significantly outperforms Qwen2.5-VL-3B by 7.8\% and achieves performance comparable to Qwen2.5-VL-72B. 
Notably, CapRL++ shows a particularly large advantage on the Coverage metric, exceeding Qwen2.5-VL-3B and Qwen2.5-VL-72B by 14.4\% and 4.0\%, respectively. 
These results highlight that our CapRL++ algorithm substantially unlocks the potential of Qwen2.5-VL-3B, enabling it to capture image details much more comprehensively.


\begin{table}[t]
    \centering
    \caption{Results on the traditional image captioning benchmark HAT \cite{petryk2024aloha}, evaluated with specialized metrics.}
    \vspace{-1mm}
    \label{tab:hat_case}
    \small
    \begin{tabular}{lccccc}
    \toprule
    Model & ALOHa & CLAIR & Factuality & Coverage & Average \\
    \midrule
    Qwen2.5-VL-3B  & 62.2 & 68.9 & 79.3 & 59.3 & 67.4 \\
    Qwen2.5-VL-72B & 66.1 & 80.6 & 83.8 & 69.7 & 75.1 \\
    $+$ CapRL++    & 65.7 & 80.7 & 80.5 & 73.7 & 75.2 \\
    \bottomrule
    \end{tabular}
    \vspace{-5mm}
\end{table}

\subsection{Video Experiments}
\label{subsec:video_experiments}

We next evaluate CapRL++ in the video domain from two complementary perspectives: whether CapRL++ captions improve video continual pretraining, and whether the trained captioner produces more informative and temporally grounded video descriptions.
Compared with static images, videos require captions to capture not only spatial content but also event ordering, state transitions, and fine-grained temporal localization.

\subsubsection{Downstream Utility via Video Continual Pretraining}
\label{subsubsec:video_pretraining}
As in the image setting, we use video-caption pairs generated by CapRL++ for continual pretraining of LVLMs.

\noindent \textbf{Implementation details.}
In our setup, we initialize the models from the Molmo2-4B \cite{clark2026molmo2} and Molmo2-8B \cite{clark2026molmo2} pretraining checkpoints to maintain training efficiency. 
The continual pretraining corpus is constructed by mixing four data types with a fixed sampling ratio of $15\% : 25\% : 30\% : 30\%$ for NLP, Image-QA, Video-QA, and Video-Caption, respectively. 
Specifically, the NLP and Image-QA data are sourced from Open-LLaVA-NeXT \cite{chen2024open}, while the video data are derived from LLaVA-Video-178K \cite{zhang2024llava}. 
To rigorously evaluate the impact of caption quality, the sole variable across our experiments is the video caption source. 
We compare the annotations generated by Qwen3-VL-4B trained with CapRL++ against the original GPT-4o  \cite{hurst2024gpt} annotations, alongside strong video-capable baselines including ShareGPT4Video-8B \cite{chen2024sharegpt4video} and Tarsier2-7B \cite{yuan2025tarsier2}.
All four settings are trained for about 5k steps, corresponding to a total training volume of approximately 1.8M samples.
All training hyperparameters and evaluation protocols follow Molmo2 \cite{clark2026molmo2}.

\noindent \textbf{Main results.}
\Cref{tab:video_pretraining_main} shows that the choice of video-caption source has a clear impact on downstream video understanding.
Among all compared settings, CapRL++ achieves the best average score on both Molmo2-4B and Molmo2-8B, indicating that it provides the strongest overall supervision for continual video pretraining.

Because CapRL++ captions include explicit timestamps, they provide dense temporal grounding signals during pretraining. This improves performance on temporal grounding benchmarks such as ActivityNet, Charades-STA, and QVHighlights.
For instance, compared with the original GPT-4o annotations, CapRL++ improves the overall average score from 39.6 to 45.9 on Molmo2-4B, and from 43.9 to 50.2 on Molmo2-8B.
Overall, these results indicate that high-quality, temporally grounded dense captions can provide meaningful benefits for video continual pretraining, whereas generic or temporally coarse caption sources offer limited advantages.

\begin{table}[!t]
  \centering
  \footnotesize
  \caption{\textbf{Comparison of different video-caption sources for continual pretraining.}
  Results are reported as average accuracy (\%) on eight downstream video understanding benchmarks using Molmo2-4B and Molmo2-8B backbones.
  CapRL++ achieves the best overall average performance across both backbones.
  \textbf{Bold} and \underline{underlined} denote the best and second-best results, respectively.}
  \vspace{0mm}
  \label{tab:video_pretraining_main}
  \begingroup
  \setlength{\tabcolsep}{4.5pt}
  \renewcommand{\arraystretch}{0.95}
  \begin{tabular}{lccccccccc}
    \toprule
    \makecell[c]{Pretraining \\ Dataset} &
    \makecell[c]{Tomato} &
    \makecell[c]{Temp\\Compass} &
    \makecell[c]{Motion\\Bench} &
    \makecell[c]{Video-\\MME} &
    \makecell[c]{LongVideo\\Bench} &
    \makecell[c]{Activity\\Net} &
    \makecell[c]{Charades-\\STA} &
    \makecell[c]{QV\\Highlights} &
    \makecell[c]{Average} \\
    \midrule
    \rowcolor{gray!10}
    \multicolumn{10}{l}{\textit{Molmo2-4B}}\\
    GPT-4o-Video-178K & 27.1 & 69.2 & 53.2 & 68.7 & \underline{61.5} & 10.2 & 14.8 & 12.4 & 39.6 \\[1.5pt]
    ShareGPT4Video-8B-Video-178K & 28.0 & 68.0 & 52.7 & 68.4 & 60.3 & 10.7 & \underline{19.3} & \underline{13.3} & 40.1 \\[1.5pt]
    Tarsier2-7B-Video-178K & \textbf{30.7} & \underline{69.9} & \underline{54.5} & \underline{69.7} & 61.0 & \underline{11.2} & 16.5 & 11.2 & \underline{40.6} \\[1.5pt]
    \rowcolor[HTML]{DAEFF9}
    CapRL-Video-178K & \textbf{30.7} & \textbf{70.1} & \textbf{55.9} & \textbf{70.2} & \textbf{62.8} & \textbf{23.1} & \textbf{31.4} & \textbf{23.1} & \textbf{45.9} \\
    \midrule
    \rowcolor{gray!10}
    \multicolumn{10}{l}{\textit{Molmo2-8B}}\\
    GPT-4o-Video-178K & 29.1 & \underline{70.4} & \underline{56.5} & 70.5 & \underline{63.3} & 17.3 & 22.9 & 21.3 & 43.9 \\[1.5pt]
    ShareGPT4Video-8B-Video-178K & 29.5 & 69.8 & 55.6 & 70.3 & 61.0 & 17.9 & 23.3 & \underline{25.3} & 44.1 \\[1.5pt]
    Tarsier2-7B-Video-178K & \textbf{31.4} & \textbf{71.8} & 55.3 & \underline{70.9} & 62.8 & \underline{18.7} & \underline{24.0} & 25.0 & \underline{45.0} \\[1.5pt]
    \rowcolor[HTML]{DAEFF9}
    CapRL-Video-178K & \underline{30.3} & \textbf{71.8} & \textbf{58.0} & \textbf{71.1} & \textbf{64.3} & \textbf{29.8} & \textbf{40.2} & \textbf{35.8} & \textbf{50.2} \\
    \bottomrule
  \end{tabular}
  \endgroup
  \vspace{-2mm}
\end{table}

\subsubsection{Intrinsic Evaluation on Video Captioning Benchmarks}
\label{subsubsec:video_caption_evaluation}

In this section, we directly evaluate the informativeness of video captions produced by CapRL++ through downstream benchmarks.

\noindent \textbf{Evaluation via the Prism framework.}
We extend the Prism evaluation to video understanding on several benchmarks.
As shown in \cref{tab:Video Prism main table}, starting from the Qwen3-VL-4B backbone, 
CapRL++ achieves an average score of 47.5\%, surpassing even Qwen3-VL-32B (46.9\%) and Qwen3-VL-235B-A22B (46.0\%).

\begin{table}[!t]
  \centering
  \footnotesize
  \caption{\small\textbf{Video captioning ability comparison in the Prism Framework.}
  Starting from Qwen3-VL-4B, CapRL++ achieves the highest average score (47.5\%), surpassing both general VLMs up to 235B scale and specialized video models.
  Notably, while TimeLens-8B~\cite{zhang2025timelens}, a Qwen3-VL-8B model specifically fine-tuned for temporal grounding, excels on TimeLens-Bench, it underperforms on general video understanding benchmarks, resulting in a lower overall score.
  In contrast, CapRL++ achieves strong performance across both temporal and general benchmarks.
  All Qwen3-VL series models used in this comparison are \textbf{Instruct} variants.
  The best results are \textbf{bold} and the second-best results are \underline{underlined}.}
  \vspace{0mm}
  \label{tab:Video Prism main table}
  \begingroup
  \setlength{\tabcolsep}{9pt}
  \renewcommand{\arraystretch}{0.85}
  \begin{tabular}{lccccccc}
    \toprule
    \multirow{2}{*}{\makecell[c]{Caption Model}}
    & \multirow{2}{*}{\makecell[c]{Video-MME}}
    & \multirow{2}{*}{\makecell[c]{MMVU}}
    & \multirow{2}{*}{\makecell[c]{MVBench}}
    & \multicolumn{3}{c}{\makecell[c]{TimeLens-Bench}}
    & \multirow{2}{*}{\makecell[c]{Average}} \\
    \cmidrule(lr){5-7}
    & & & & Charades & ActivityNet & QVHighlights & \\
    \midrule
    \rowcolor{gray!10}
    \multicolumn{8}{l}{\textit{General VLMs}} \\
    Qwen3-VL-4B (\textit{baseline})     & 51.8 & 45.7 & 56.8 & 14.9 & 8.9  & 7.4  & 41.2 \\[1.5pt]
    Qwen3-VL-8B                         & 54.8 & 46.9 & 58.3 & 26.7 & 11.1 & 9.3  & 43.9 \\[1.5pt]
    Qwen3-VL-32B                        & \underline{56.8} & \textbf{51.1} & \underline{60.5} & 30.9 & 13.6 & 11.1 & \underline{46.9} \\[1.5pt]
    Qwen3-VL-235B-A22B                  & 56.7 & \underline{51.0} & 59.5 & 28.0 & 12.8 & 9.9  & 46.0 \\[1.5pt]
    \midrule
    \rowcolor{gray!10}
    \multicolumn{8}{l}{\textit{Specialized Video VLMs}} \\
    Tarsier2-7B \cite{yuan2025tarsier2}  & 49.7 & 43.3 & 49.7 & 9.0 & 8.7 & 8.1 & 37.8 \\[1.5pt]
    TimeLens-8B \cite{zhang2025timelens} & 53.8 & 47.7 & 57.8 & \underline{31.0} & \textbf{23.4} & \textbf{23.1} & 46.3 \\[1.5pt]
    \midrule
    \rowcolor[HTML]{EAF4EA}
    Qwen3-VL-as-Judge-4B                & 56.5 & 47.1 & 57.0 & 27.6 & 14.1 & 13.1 & 44.7 \\[1.5pt]
    \rowcolor[HTML]{DAEFF9}
    Ours (w/ CapRL++)                   & \textbf{59.4} & 48.5 & \textbf{60.7} & \textbf{34.3} & \underline{15.6} & \underline{14.2} & \textbf{47.5} \\
    \bottomrule
  \end{tabular}
  \endgroup
  \vspace{-1mm}
\end{table}

Consistent with the video pretraining results, the improvements are most pronounced on \textbf{temporal grounding} tasks. 
On TimeLens-Bench, which tests fine-grained temporal localization, standard LVLMs struggle significantly (e.g., 10.4 for Qwen3-VL-4B and 16.9 for 235B) due to their tendency to generate temporally vague descriptions. 
In contrast, CapRL++ achieves an average score of 21.4. While ranking second only to the task-specific TimeLens-8B \cite{zhang2025timelens} model, it more than doubles the 4B baseline (+106\% relative improvement) and surpasses the massive 235B model by 4.5 points. 
This strong temporal capability is further evidenced by the Video-MME temporal perception subtask, where CapRL++ achieves 62.3\% (+14.6 points over the baseline).

We attribute these gains to the temporal localization MCQs in our QA-driven reward. Unlike conventional SFT objectives, this reward penalizes captions that omit event boundaries and timestamps, encouraging the policy model to encode chronological cues that improve QA accuracy.

Ultimately, these results highlight that \textbf{temporal perception constitutes the key new dimension} in video captioning. 
While spatial understanding can largely transfer zero-shot from image-level training (as shown in \cref{tab:cross_modal}), fine-grained temporal reasoning demands explicit optimization. 
CapRL++ effectively bridges this gap, establishing a unified paradigm that improves both spatial and temporal visual understanding.

\noindent \textbf{Evaluation on more video caption benchmarks.}
We further evaluate caption comprehensiveness on Dream-1K \cite{wang2024tarsier} and CaReBench \cite{xu2024carebench}. 
Since the original Dream-1K references are brief and can penalize dense but valid predictions \cite{xu2024carebench}, we report results under a re-annotated evaluation setting, where Gemini-3-Flash provides denser references and matching is relaxed for visually plausible details absent from the original references.

As shown in \cref{tab:dream_carebench}, CapRL++ achieves the highest \textbf{F1} scores across all three benchmarks.
The gains in \textbf{Recall} are particularly notable: despite using the Qwen3-VL-4B backbone, CapRL++ exceeds Qwen3-VL-235B-A22B by +5.9 on Dream-1K, +14.0 on CaReBench Action, and +22.4 on CaReBench Object, suggesting stronger coverage of video details.

The lower \textbf{Precision} reflects a natural trade-off in generating substantially denser captions. Greater verbosity increases the chance of minor deviations and occasional hallucinations, while also enabling the model to cover many additional valid visual details beyond those described in the references.

\begin{table}[!t]
  \centering
  \caption{\textbf{LLM-as-a-judge scores on Dream-1K and CaReBench.}
  Ground-truth captions on Dream-1K are re-annotated with Gemini-3-Flash to provide denser and more detailed references.
  CapRL++ achieves the highest \textbf{F1} and \textbf{Recall} across all benchmarks, demonstrating its superior ability to capture video content comprehensively.
  Its relatively lower \textbf{Precision} reflects the inherent trade-off of generating substantially longer and denser captions: the increased verbosity naturally introduces a higher likelihood of minor deviations, while also producing many valid visual details that exceed the scope of the shorter references, making precise reference alignment more challenging. The best results are \textbf{bold} and the second-best results are \underline{underlined}.}
  \label{tab:dream_carebench}
  \renewcommand{\arraystretch}{1.08}
  \footnotesize
  \begin{tabular*}{\columnwidth}{@{\extracolsep{\fill}} l|ccc|ccc|ccc|c}
    \toprule
    \multicolumn{1}{c|}{\multirow{2}{*}{\makecell[c]{Caption\\Model}}}
    & \multicolumn{3}{c|}{Dream-1K}
    & \multicolumn{3}{c|}{CaReBench-Action}
    & \multicolumn{3}{c|}{CaReBench-Object}
    & \multirow{2}{*}{Avg. F1} \\
    \cmidrule(lr){2-4}\cmidrule(lr){5-7}\cmidrule(lr){8-10}
    & Recall & Precision & F1
    & Recall & Precision & F1
    & Recall & Precision & F1
    &  \\
    \midrule
    Qwen3-VL-4B (\textit{baseline})
    & 33.4 & \underline{89.4} & 48.6
    & 23.9 & \underline{81.5} & 36.9
    & 29.9 & \textbf{80.3} & 43.6
    & 43.0 \\
    Qwen3-VL-8B
    & 35.8 & 88.9 & 51.1
    & 26.1 & \textbf{82.1} & 39.6
    & 32.2 & \underline{79.2} & 45.8
    & 45.5 \\
    Qwen3-VL-235B-A22B
    & \underline{39.5} & \textbf{89.5} & \underline{54.8}
    & \underline{26.6} & 81.4 & \underline{40.1}
    & \underline{32.4} & \textbf{80.3} & \underline{46.1}
    & \underline{47.0} \\
    Ours (w/ CapRL++)
    & \textbf{45.4} & 82.3 & \textbf{58.5}
    & \textbf{40.6} & 70.5 & \textbf{51.5}
    & \textbf{54.8} & 70.6 & \textbf{61.7}
    & \textbf{57.2} \\
    \bottomrule
  \end{tabular*}
  \vspace{-2mm}
\end{table}

\subsection{Discussion and Ablation Studies}
\label{subsec:discussion_and_ablations}

In this section, we provide a comprehensive analysis and discussion of CapRL++, covering both image and video modalities.
These results further confirm CapRL++'s general applicability, robustness, and effectiveness.

\begin{table*}[!t]
  \centering
  \caption{\textbf{Cross-modal captioning performance of CapRL++.}
  We compare three training paradigms: Image-only, Video-only, and SpaBoot (Image$\rightarrow$Video), all starting from Qwen3-VL-4B,
  and evaluate their transfer behavior on both image and video benchmarks via the Prism framework.
  The SpaBoot paradigm achieves the best overall balance,
  retaining strong image captioning performance while maximizing video understanding.
  The best results are \textbf{bold} and the second-best results are \underline{underlined}.}
  \label{tab:cross_modal}
  \resizebox{\textwidth}{!}{%
  \begin{tabular}{l c l c c c c c c c c c c}
    \toprule
    \multicolumn{3}{c}{\multirow{2}{*}[-6pt]{Model}}
      & \multicolumn{6}{c}{\textit{Image Benchmarks}}
      & \multicolumn{4}{c}{\textit{Video Benchmarks}} \\
    \cmidrule(lr){4-9}\cmidrule(lr){10-13}
    \multicolumn{3}{c}{}
    & ChartQA & MMStar & MMMU-Pro & CharXiv & SEED & Avg
    & MMVU & QVHighlights & ActivityNet & Avg \\
    \midrule
    \multirow{3}{*}{Qwen3-VL}
    & {\vrule width 0.6pt}
    & 4B (\textit{baseline})  & 41.1 & 58.0 & 33.3 & 32.4 & 72.0 & 47.4 & 45.7 & 7.4 & 8.9 & 20.7 \\
    & {\vrule width 0.6pt}
    & 32B                     & \underline{43.5} & 60.7 & \underline{35.5} & \underline{38.3} & 72.6 & 50.1 & \textbf{51.1} & 11.0 & 13.6 & \underline{25.2} \\
    & {\vrule width 0.6pt}
    & 235B-A22B               & \textbf{44.9} & \textbf{61.3} & \textbf{36.3} & 37.5 & 72.6 & \textbf{50.5} & \underline{51.0} & 9.9 & 12.8 & 24.6 \\
    \midrule
    \multirow{3}{*}{$+$ CapRL++}
    & {\vrule width 0.6pt}
    & Image-only              & 41.2 & 59.7 & 33.9 & 38.2 & \underline{73.3} & 49.3 & 46.0 & 8.9 & 10.3 & 21.7 \\
    & {\vrule width 0.6pt}
    & Video-only              & 42.8 & 60.6 & 33.4 & 36.3 & 73.2 & 49.3 & 47.5 & \underline{13.3} & \underline{14.1} & 25.0 \\
    & {\vrule width 0.6pt}
    & SpaBoot              & 41.9 & \underline{61.2} & 34.4 & \textbf{39.9} & \textbf{74.1} & \underline{50.3} & 48.5 & \textbf{14.2} & \textbf{15.6} & \textbf{26.1} \\
    \bottomrule
  \end{tabular}%
  }
  \vspace{-1mm}
\end{table*}

\noindent \textbf{Cross-modal transferability and SpaBoot evaluation.} 
To quantitatively validate our proposed SpaBoot strategy (\cref{subsec:curriculum_method}), we evaluate three training paradigms: Image-only, Video-only, and SpaBoot, across both image and video benchmarks (\cref{tab:cross_modal}). 

As shown in \cref{tab:cross_modal}, the \textbf{SpaBoot} paradigm emerges as the optimal configuration. On video benchmarks, it consistently surpasses the Video-only approach (26.1 vs. 25.0 avg.), even outperforming the much larger Qwen3-VL-32B and 235B models. On image benchmarks, the SpaBoot model retains near-upper-bound spatial performance (50.3 avg.), exceeding the Image-only model. This confirms that our SpaBoot achieves a Pareto-optimal trade-off, maximizing temporal comprehension without degrading spatial awareness. 

In contrast, while \textbf{Video-only} training yields strong video gains, its performance on fine-grained spatial reasoning tasks (e.g., CharXiv) is visibly inferior, validating our hypothesis that video frames are a less efficient source for static spatial alignment. We further observe positive \textbf{bidirectional transfer} between modalities: \textbf{Image-only} training yields zero-shot gains on video benchmarks (21.7 vs.\ 20.7 avg.), and conversely \textbf{Video-only} training improves image-side performance (49.3 vs.\ 47.4 avg.), suggesting that the utility reward captures transferable visual representations rather than modality-specific cues. These quantitative results are consistent with the qualitative progression illustrated previously in \cref{fig:case_study}.

\begin{figure*}[!t]
  \centering
  \includegraphics[width=0.99\textwidth]{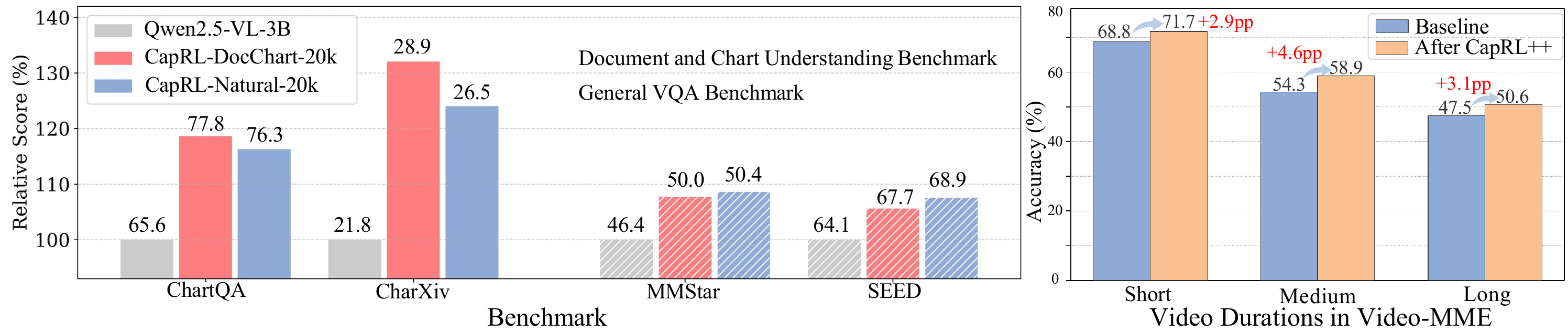}
  \vspace{-1mm}
  \caption{\textbf{(Left)} CapRL++ demonstrates strong generalization even when trained on images from a single domain. 
  CapRL-DocChart-20k refers to training conducted solely on document or chart images, while CapRL-Natural-20k is trained exclusively on natural images. 
  Both models achieve significant improvements over the baseline on out-of-domain benchmarks, highlighting strong generalization capability.
  \textbf{(Right)} CapRL++ also exhibits strong generalization across video durations. 
  Notably, despite being trained exclusively on short video clips under 30 seconds, CapRL++ achieves substantial improvements over the baseline on VideoMME across all sequence lengths, including Short ($<2$ min), Medium (4--15 min), and Long (30--60 min).}
  \vspace{-5mm}
  \label{fig:domain_and_duration_generalization}
\end{figure*}

\noindent \textbf{Qualitative analysis of SpaBoot dynamics.} 
To complement the quantitative results, \cref{fig:case_study} visualizes the progressive capability gains across the SpaBoot stages using a tug-of-war case study. 
The original model produces only coarse descriptions of the scene.
Although it roughly captures the indoor arena and the tug-of-war setting, its captions lack fine-grained spatial/object grounding and provide limited event-level details.
After image-stage CapRL++ training, the model demonstrates markedly enhanced perceptual grounding (e.g., precise OCR and fine-grained attribute extraction). However, its temporal grounding remains insufficient: the descriptions fail to precisely align evolving actions with the corresponding timestamps.
Finally, the full SpaBoot model not only retains the spatial acuity acquired during the image stage but also introduces well-structured narrative organization and precise temporal segmentation. 
This progression qualitatively corroborates our core finding: SpaBoot first strengthens perceptual grounding through image supervision and then builds temporal understanding on top of these spatial representations, leading to more complete and grounded video captions.

\noindent \textbf{CapRL++ exhibits strong generalization across image domains and video durations.}
We first investigate the effect of different image sources in the training data. To this end, we classify the images into two categories using Qwen2.5-VL-3B: (1) documents, charts, or infographics, and (2) natural images. 
From each category, we sample 20k images for comparison.
As illustrated in \cref{fig:domain_and_duration_generalization} (Left), models trained exclusively on chart-type images via GRPO exhibit substantial gains over Qwen2.5-VL-3B, not only in document and chart understanding but also in general VQA tasks. 


\begin{table}[t]
  \centering
  \captionsetup{justification=centering}
  \caption{\small\textbf{Ablations about Sampling Rounds $N$.}}
  \vspace{-3mm}
  \label{tab:Ablations about N}
  \footnotesize
  \renewcommand{\arraystretch}{1.0}
  \begin{tabular*}{\linewidth}{@{\extracolsep{\fill}}lcccccccccc}
    \toprule
    \multirow{2}{*}[-1ex]{\makecell{Sampling\\Rounds}}
    & \multicolumn{6}{c}{\textit{Image CapRL++}}
    & \multicolumn{4}{c}{\textit{Video CapRL++}} \\
    \cmidrule(lr){2-7} \cmidrule(lr){8-11}
    & \makecell{ChartQA\\Pro} & \makecell{Info\\VQA} & MMMU & MMStar & WeMath & Avg
    & \makecell{Video-MME} & MVBench & \makecell{TimeLens\\-Bench} & Avg \\
    \midrule
    N=1 & 35.4 & 58.1 & 36.5 & 50.2 & 56.1 & 47.3 & 55.8 & 59.5 & 16.5 & 43.9 \\
    N=2 & 36.2 & 59.1 & 36.3 & 49.3 & 56.9 & 47.6 & 58.5 & 60.4 & 18.3 & 45.7 \\
    N=4 & 36.7 & 59.9 & 37.1 & 50.9 & 57.3 & 48.4 & 58.0 & 60.4 & 19.5 & 46.0 \\
    N=8 & 36.9 & 59.6 & 36.5 & 50.8 & 57.7 & 48.3 & 58.3 & 60.2 & 18.6 & 45.7 \\
    \bottomrule
  \end{tabular*}
  \vspace{-2mm}
\end{table}

Beyond image domains, we observe a similarly strong generalization across temporal lengths in video understanding.
As shown in \cref{fig:domain_and_duration_generalization} (Right), CapRL++ demonstrates significant performance improvements even on long videos (30--60 min),
despite the training phase primarily utilizing short video clips ($<30$\,s).
This is particularly noteworthy because longer videos demand more sophisticated temporal reasoning: the model must track events, state changes, and causal relationships over extended time spans.
The fact that CapRL++ generalizes to these challenging settings suggests that the temporal perception skills acquired through QA-driven rewards are not superficial pattern matching on short clips, but reflect a deeper understanding of temporal structure that transfers across durations.
Together, these findings indicate that the detailed perception and reasoning capabilities unlocked by CapRL++ are highly transferable, generalizing robustly beyond the specific visual domains and temporal formats encountered during training.

\begin{figure*}[t!]
  \centering
  \includegraphics[width=1\textwidth]{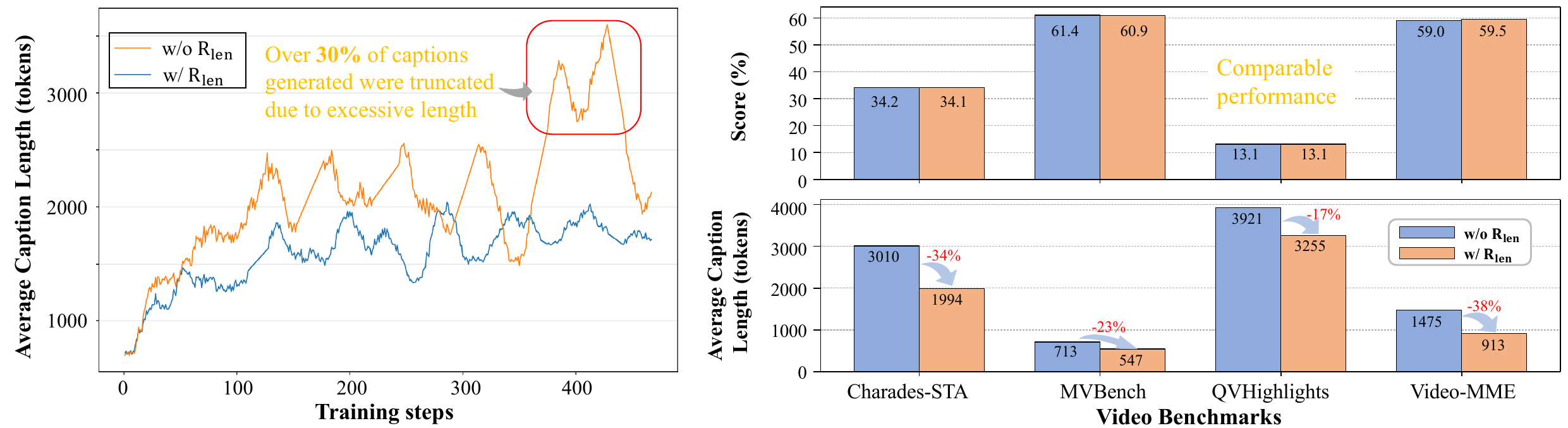}
  \vspace{-15pt}
  \caption{
  \textbf{(Left)} As training progresses, $R_{\text{len}}$ helps the length of model-generated captions converge to a stable range, 
  whereas without it, we observe that over 30\% of responses are truncated due to excessive length in our training process.
  \textbf{(Right)} The captioner trained with $R_{\text{len}}$ achieves comparable performance across multiple video benchmarks with more concise captions.
  } 
  \vspace{-8pt}
  \label{fig:ablation_r_len}
\end{figure*}

\begin{figure}[t!]
  \centering
  \includegraphics[width=0.65\columnwidth]{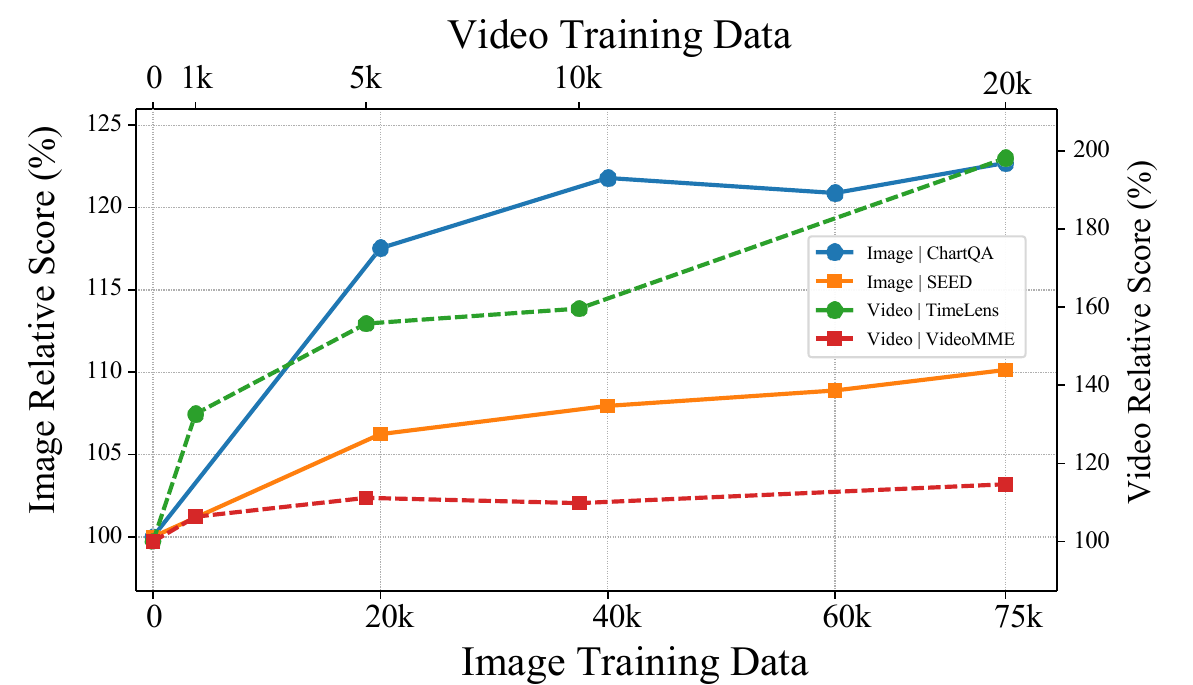}
  \vspace{-5pt}
  \caption{\small\textbf{Scaling behavior of CapRL++ with increasing QA training data.}
  Both image and video tasks exhibit promising scaling trends as more QA training samples are used.
  Notably, video tasks achieve substantial gains with relatively fewer data, reflecting the higher information density per sample.
  On video benchmarks, general-purpose evaluation (e.g., VideoMME) shows only moderate improvement with additional data, whereas temporal understanding (TimeLens) benefits dramatically, reaching roughly $2\times$ the initial score at 20k samples.}
  \label{fig:scaling_trend}
  \vspace{-2mm}
\end{figure}

\noindent \textbf{CapRL++ demonstrates promising scaling behavior with training data.}
We study the scaling behavior of CapRL++ by varying the amount of image and video QA training data.
As shown in \cref{fig:scaling_trend}, both image and video tasks exhibit clear and encouraging scaling trends: model performance improves steadily as the volume of QA supervision increases.
Interestingly, the two modalities differ in data efficiency.
For image tasks, performance scales smoothly and continuously with additional QA data, indicating that CapRL++ can progressively unlock the visual understanding potential of the base model as more image QA supervision becomes available.
For video tasks, the model achieves considerable gains with relatively fewer training samples, which we attribute to the inherently higher information density of videos—each sample contains a large number of visual frames along with rich temporal and semantic cues, providing denser supervision per instance.

A closer look at the video results reveals an intriguing divergence across evaluation dimensions.
On general video understanding benchmarks such as VideoMME, performance improves only moderately as the training data scales up, suggesting that broad video understanding is already reasonably well captured with a modest amount of QA data.
In stark contrast, temporal understanding—as measured by TimeLens-Bench—benefits dramatically from increased data, with the score at 20k training samples reaching approximately $2\times$ that of the smallest data setting.
This indicates that fine-grained temporal reasoning is a capability that scales particularly well with additional QA supervision under the CapRL++ framework, and that the training data primarily unlocks deeper temporal comprehension rather than general visual recognition.
Given the relatively small parameter sizes of the base models used and the favorable scaling behavior observed across both modalities, CapRL++ holds considerable promise for larger-scale multimodal training.

\noindent \textbf{Sparse QA supervision is sufficient for CapRL++.} 
We examine the effect of varying the number of QA pairs per image and per video. 
For images, we randomly sample 20k images that retain three QA pairs after filtering, and train separate models using 1, 2, or 3 QA pairs per image.
For videos, we sample 10k videos and similarly vary the number of QA pairs per video among 1, 4, and 8.
The results are presented in \cref{tab:Analysis_about_number_of_QA_numbers}.
For images, even a single QA pair per image yields a substantial improvement, raising the average score from 40.6 to 48.0 (+7.4), which accounts for the vast majority of the total gain; increasing to 2 or 3 QA pairs provides only marginal further improvement (+0.5).
A similar pattern emerges for videos: using just 1 QA pair per video already boosts the average from 39.7 to 45.2 (+5.5), while scaling up to 4 or 8 QA pairs brings diminishing returns.
These results highlight the remarkable data efficiency of CapRL++: highly sparse QA supervision is sufficient to unlock the majority of captioning gains, making the approach practical even when dense QA annotations are unavailable.

\begin{table}[t]
  \centering
  \captionsetup{justification=centering}
  \caption{\small\textbf{Analysis of the number of QA per image and video.}}
  \vspace{-2mm}
  \label{tab:Analysis_about_number_of_QA_numbers}
  \footnotesize
  \renewcommand{\arraystretch}{1.0}
  \setlength{\tabcolsep}{4pt}
  \resizebox{\columnwidth}{!}{%
  \begin{tabular}{ccccccc@{\hspace{8mm}}ccccc}
    \toprule
    \multicolumn{7}{c}{\textit{Image CapRL++}} & \multicolumn{5}{c}{\textit{Video CapRL++}} \\
    \cmidrule(r){1-7} \cmidrule(l){8-12}
    \makecell{\#QA/image} & \makecell{ChartQA\\Pro} & \makecell{Info\\VQA} & MMMU & MMStar & WeMath & Avg
    & \makecell{\#QA/video} & \makecell{Video-MME} & MVBench & \makecell{TimeLens\\-Bench} & Avg \\
    \midrule
    \textit{baseline} & 27.1 & 40.2 & 35.1 & 46.4 & 54.4 & 40.6
    & \textit{baseline} & 51.8 & 56.8 & 10.4 & 39.7 \\
    1                 & 35.5 & 59.8 & 36.6 & 50.8 & 57.3 & 48.0
    & 1                 & 58.2 & 60.3 & 17.0 & 45.2 \\
    2                 & 36.8 & 60.2 & 37.6 & 51.1 & 56.6 & 48.5
    & 4                 & 58.8 & 61.4 & 19.4 & 46.5 \\
    3                 & 36.9 & 60.3 & 36.9 & 51.3 & 56.8 & 48.5
    & 8                 & 59.2 & 60.6 & 19.8 & 46.5 \\
    \bottomrule
  \end{tabular}%
  }
  \vspace{-2mm}
\end{table}

\noindent \textbf{Ablations about sampling rounds $N$.}
We investigate the effect of the number of sampling rounds $N$ used for computing the QA-based reward.
As shown in \cref{tab:Ablations about N}, performance across both modalities improves significantly with multiple sampling rounds, reaching a stable plateau between $N=4$ and $N=8$. 
Specifically, compared to $N=1$, the average score at $N=8$ increases from 47.3 to 48.3 for images, and jumps from 43.9 to 45.7 for videos. 
The poor performance at $N=1$ stems from the vision-free LLM's inherent option-position biases; without sufficient option shuffling, single-round evaluation yields a noisy reward that misdirects policy optimization. 
Increasing the sampling rounds effectively marginalizes out these biases. 
To ensure maximum stability and minimal variance in the reward signal during training, we adopt $N=8$ as the default configuration for CapRL++.

\noindent \textbf{Ablations about length reward $R_{\text{len}}$.}
As shown in \cref{fig:ablation_r_len} (Left), without length regularization, 
the model tends to generate increasingly longer captions as training progresses 
to cover a broader range of information, thereby improving the accuracy of the 
vision-free model in answering MCQs. However, these longer captions often contain 
substantial redundant information, which also undermines training efficiency. 
As shown in \cref{fig:ablation_r_len} (Right), $R_{\text{len}}$ helps strike a 
balance between caption length and informativeness, enabling the model to convey 
more useful information within an appropriate length and ultimately achieve 
better caption quality. Moreover, the shorter outputs induced by $R_{\text{len}}$ 
yield a practical efficiency benefit: we observe approximately 9\% speedup in 
rollout throughput compared to training without $R_{\text{len}}$, further 
improving the overall training efficiency of CapRL++.


\begin{table}[t]
  \centering
  \caption{\small\textbf{Text-to-image generation performance on UniGenBench.}
  CapRL-T2I improves instruction-following ability on both FLUX.1-dev and FLUX2 Klein 9B.}
  \vspace{-1mm}
  \label{tab:unigenbench_flux_caprl_reverse}
  \setlength{\tabcolsep}{2pt}
  \renewcommand{\arraystretch}{0.85}
  \scalebox{0.9}{
  \begin{tabular}{lccccc}
    \toprule
    Model & \makecell{Text\\Generation} & Attribute & Action & \makecell{Logical\\Reasoning} & Overall \\
    \midrule
    FLUX.1-dev         & 34.48 & 64.32 & 61.12 & 27.52 & 59.92 \\
    $+$ CapRL-T2I      & \textbf{50.29} & \textbf{79.49} & \textbf{75.38} & \textbf{44.50} & \textbf{71.77} \\
    \midrule
    FLUX2 Klein 9B     & 55.33 & 82.91 & 76.52 & 55.05 & 78.59 \\
    $+$ CapRL-T2I      & \textbf{57.39} & \textbf{85.04} & \textbf{79.18} & \textbf{61.24} & \textbf{80.86} \\
    \bottomrule
  \end{tabular}}
  \vspace{-2mm}
\end{table}


\noindent \textbf{Beyond captioning: Extending utility-based rewards to text-to-image generation.}
The utility-as-reward principle is not tied to captioning.
In CapRL++, a caption is rewarded by how well it enables a vision-free LLM to answer questions about the source image.
We extend this idea to text-to-image (T2I) generation by reversing the direction of the proxy: a generated image is rewarded if a VLM can answer prompt-grounded questions from the image alone.
We instantiate this inversion as CapRL-T2I: given a text prompt, we construct prompt-grounded MCQs offline and obtain images from the T2I model during RL. A vision-grounded VLM then answers these MCQs from each generated image alone, and its accuracy serves as the reward for that image.
This accuracy measures how faithfully the generated image realizes the prompt and serves as a verifiable reward for reinforcement fine-tuning of the generator.

\begin{figure}[t]
  \centering
  \includegraphics[width=0.65\columnwidth]{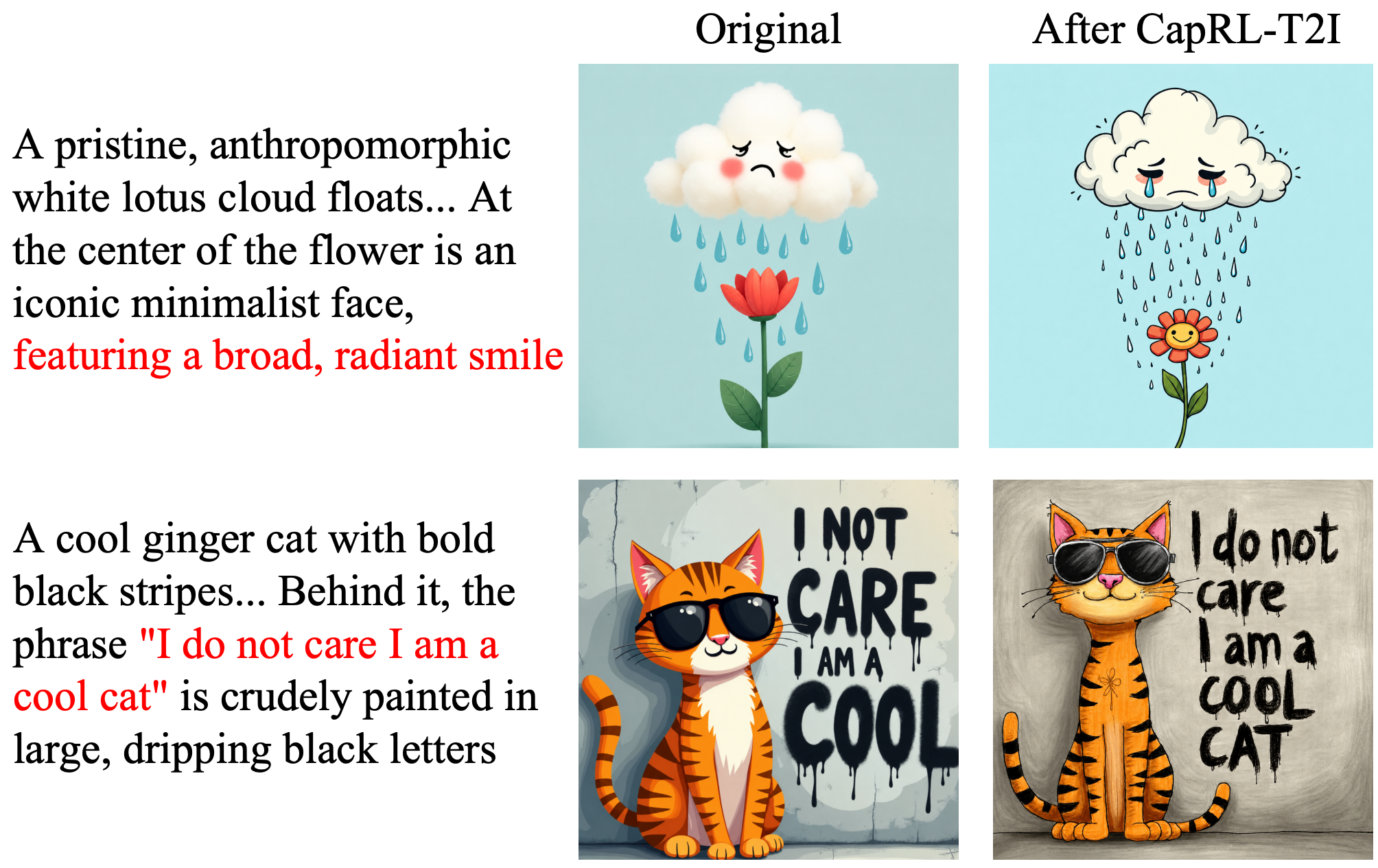}
  \caption{\textbf{Qualitative comparison} of CapRL-T2I on FLUX.1-dev. CapRL-T2I yields stronger instruction-following ability.}
  \label{fig:CapRL-T2I_01}
  \vspace{-4mm}
\end{figure}

We build training data from the prompts of UniGenBench~\cite{Pref-GRPO&UniGenBench}.
We expand each prompt to around 100 tokens with Gemini-3-Flash to raise the instruction-following difficulty. Following the QA construction in CapRL++, we generate prompt-grounded MCQs with Gemini-3-Flash and filter them with Qwen3-8B, keeping only questions answered correctly given the prompt and incorrectly without it.
We reinforce the T2I model on the DanceGRPO~\cite{xue2025dancegrpo} codebase, using Qwen3-VL-8B as the vision-grounded answerer.
As shown in \cref{tab:unigenbench_flux_caprl_reverse}, CapRL-T2I improves the overall UniGenBench score on both backbones, from 59.92 to 71.77 on FLUX.1-dev \cite{flux2024} and from 78.59 to 80.86 on FLUX2 Klein 9B \cite{flux-2-2025}, where the smaller gain on the latter is consistent with its higher baseline.
The improvement is largest on text generation (34.48 to 50.29 on FLUX.1-dev), indicating that the MCQ-based reward effectively penalizes prompt elements the generated image fails to realize.
The qualitative results in \cref{fig:CapRL-T2I_01} further show improved prompt following, supporting the broader applicability of utility-based rewards beyond captioning.

\section{Conclusion}
\label{sec:conclusion}

In this paper, we present CapRL++, a unified reinforcement learning framework that redefines multimodal caption quality through verifiable information utility rather than reference imitation. By optimizing for the accuracy of a vision-free model in answering questions based on generated captions, CapRL++ effectively bypasses the high annotation costs and subjective biases inherent in traditional supervised fine-tuning. Through comprehensive evaluations on more than 20 benchmarks across image and video settings, we demonstrate that CapRL++ enables compact models to achieve exceptional dense captioning performance, comparable to and in some cases surpassing models with significantly larger parameter counts. Furthermore, extensive ablation studies validate our core designs, including reward formulation, Spatial-Anchored Bootstrapping (SpaBoot) strategy, and data scaling, confirming that CapRL++ exhibits robust generalization and remarkable data efficiency. Ultimately, this work establishes utility-based verifiable rewards as a scalable and robust foundation for advancing the next generation of vision-language models.

\FloatBarrier

\clearpage
\bibliographystyle{plain}
\bibliography{refs}

@article{shao2024deepseekmath,
  title={{DeepseekMath}: Pushing the limits of mathematical reasoning in open language models},
  author={Shao, Zhihong and Wang, Peiyi and Zhu, Qihao and Xu, Runxin and Song, Junxiao and Bi, Xiao and Zhang, Haowei and Zhang, Mingchuan and Li, YK and Wu, Yang and others},
  journal={arXiv preprint arXiv:2402.03300},
  year={2024}
}

@inproceedings{masry2022chartqa,
  title={Chartqa: A benchmark for question answering about charts with visual and logical reasoning},
  author={Masry, Ahmed and Do, Xuan Long and Tan, Jia Qing and Joty, Shafiq and Hoque, Enamul},
  booktitle={Findings of the association for computational linguistics: ACL 2022},
  pages={2263--2279},
  year={2022}
}

@inproceedings{mathew2021docvqa,
  title={Docvqa: A dataset for vqa on document images},
  author={Mathew, Minesh and Karatzas, Dimosthenis and Jawahar, CV},
  booktitle={Proceedings of the IEEE/CVF winter conference on applications of computer vision},
  pages={2200--2209},
  year={2021}
}

@inproceedings{liu2024tempcompass,
  title={Tempcompass: Do video llms really understand videos?},
  author={Liu, Yuanxin and Li, Shicheng and Liu, Yi and Wang, Yuxiang and Ren, Shuhuai and Li, Lei and Chen, Sishuo and Sun, Xu and Hou, Lu},
  booktitle={Findings of the Association for Computational Linguistics: ACL 2024},
  pages={8731--8772},
  year={2024}
}

@article{wu2024longvideobench,
  title={Longvideobench: A benchmark for long-context interleaved video-language understanding},
  author={Wu, Haoning and Li, Dongxu and Chen, Bei and Li, Junnan},
  journal={Advances in Neural Information Processing Systems},
  volume={37},
  pages={28828--28857},
  year={2024}
}

@inproceedings{hong2025motionbench,
  title={Motionbench: Benchmarking and improving fine-grained video motion understanding for vision language models},
  author={Hong, Wenyi and Cheng, Yean and Yang, Zhuoyi and Wang, Weihan and Wang, Lefan and Gu, Xiaotao and Huang, Shiyu and Dong, Yuxiao and Tang, Jie},
  booktitle={Proceedings of the Computer Vision and Pattern Recognition Conference},
  pages={8450--8460},
  year={2025}
}

@article{yu2023mm,
  title={Mm-vet: Evaluating large multimodal models for integrated capabilities},
  author={Yu, Weihao and Yang, Zhengyuan and Li, Linjie and Wang, Jianfeng and Lin, Kevin and Liu, Zicheng and Wang, Xinchao and Wang, Lijuan},
  journal={arXiv preprint arXiv:2308.02490},
  year={2023}
}

@inproceedings{shangguan2025tomato,
  title={Tomato: Assessing visual temporal reasoning capabilities in multimodal foundation models},
  author={Shangguan, Ziyao and Li, Chuhan and Ding, Yuxuan and Zheng, Yanan and Zhao, Yilun and Fitzgerald, Tesca and Cohan, Arman},
  booktitle={International Conference on Learning Representations},
  volume={2025},
  pages={7593--7734},
  year={2025}
}

@inproceedings{hudson2019gqa,
  title={Gqa: A new dataset for real-world visual reasoning and compositional question answering},
  author={Hudson, Drew A and Manning, Christopher D},
  booktitle={Proceedings of the IEEE/CVF conference on computer vision and pattern recognition},
  pages={6700--6709},
  year={2019}
}

@inproceedings{liu2024mmbench,
  title={Mmbench: Is your multi-modal model an all-around player?},
  author={Liu, Yuan and Duan, Haodong and Zhang, Yuanhan and Li, Bo and Zhang, Songyang and Zhao, Wangbo and Yuan, Yike and Wang, Jiaqi and He, Conghui and Liu, Ziwei and others},
  booktitle={European conference on computer vision},
  pages={216--233},
  year={2024},
  organization={Springer}
}

@article{li2024seed,
  title={Seed-bench-2-plus: Benchmarking multimodal large language models with text-rich visual comprehension},
  author={Li, Bohao and Ge, Yuying and Chen, Yi and Ge, Yixiao and Zhang, Ruimao and Shan, Ying},
  journal={arXiv preprint arXiv:2404.16790},
  year={2024}
}

@misc{xai_org_realworldqa,
  author       = {{xai-org}},
  title        = {RealworldQA},
  year         = {2024},
  publisher    = {Hugging Face},
  howpublished = {\url{https://huggingface.co/datasets/xai-org/RealworldQA}},
  note         = {Accessed: 2026-05-11}
}

@inproceedings{kembhavi2016diagram,
  title={A diagram is worth a dozen images},
  author={Kembhavi, Aniruddha and Salvato, Mike and Kolve, Eric and Seo, Minjoon and Hajishirzi, Hannaneh and Farhadi, Ali},
  booktitle={European conference on computer vision},
  pages={235--251},
  year={2016},
  organization={Springer}
}

@inproceedings{lu2024mathvista,
  title={Mathvista: Evaluating mathematical reasoning of foundation models in visual contexts},
  author={Lu, Pan and Bansal, Hritik and Xia, Tony and Liu, Jiacheng and Li, Chunyuan and Hajishirzi, Hannaneh and Cheng, Hao and Chang, Kai-Wei and Galley, Michel and Gao, Jianfeng},
  booktitle={International Conference on Learning Representations},
  volume={2024},
  pages={23439--23554},
  year={2024}
}

@misc{openai2023gpt4v_system_card,
  author       = {{OpenAI}},
  title        = {{GPT-4V(ision) System Card}},
  year         = {2023},
  howpublished = {\url{https://cdn.openai.com/papers/GPTV_System_Card.pdf}},
  note         = {Accessed: 2026-05-08}
}

@article{hurst2024gpt,
  title={Gpt-4o system card},
  author={Hurst, Aaron and Lerer, Adam and Goucher, Adam P and Perelman, Adam and Ramesh, Aditya and Clark, Aidan and Ostrow, AJ and Welihinda, Akila and Hayes, Alan and Radford, Alec and others},
  journal={arXiv preprint arXiv:2410.21276},
  year={2024}
}

@article{lambert2024tulu,
  title={{Tulu 3}: Pushing frontiers in open language model post-training},
  author={Lambert, Nathan and Morrison, Jacob and Pyatkin, Valentina and Huang, Shengyi and Ivison, Hamish and Brahman, Faeze and Miranda, Lester James V and Liu, Alisa and Dziri, Nouha and Lyu, Shane and others},
  journal={arXiv preprint arXiv:2411.15124},
  year={2024}
}

@article{wang2024tarsier,
  title={Tarsier: Recipes for training and evaluating large video description models},
  author={Wang, Jiawei and Yuan, Liping and Zhang, Yuchen and Sun, Haomiao},
  journal={arXiv preprint arXiv:2407.00634},
  year={2024}
}

@article{yuan2025tarsier2,
  title={Tarsier2: Advancing large vision-language models from detailed video description to comprehensive video understanding},
  author={Yuan, Liping and Wang, Jiawei and Sun, Haomiao and Zhang, Yuchen and Lin, Yuan},
  journal={arXiv preprint arXiv:2501.07888},
  year={2025}
}

@article{zhang2024llava,
  title={Llava-video: Video instruction tuning with synthetic data},
  author={Zhang, Yuanhan and Wu, Jinming and Li, Wei and Li, Bo and Ma, Zejun and Liu, Ziwei and Li, Chunyuan},
  journal={arXiv preprint arXiv:2410.02713},
  year={2024}
}

@article{xu2024carebench,
  title={Carebench: A fine-grained benchmark for video captioning and retrieval},
  author={Xu, Yifan and Li, Xinhao and Yang, Yichun and Meng, Desen and Huang, Rui and Wang, Limin},
  journal={arXiv preprint arXiv:2501.00513},
  year={2024}
}

@inproceedings{fu2025video,
  title={Video-mme: The first-ever comprehensive evaluation benchmark of multi-modal llms in video analysis},
  author={Fu, Chaoyou and Dai, Yuhan and Luo, Yongdong and Li, Lei and Ren, Shuhuai and Zhang, Renrui and Wang, Zihan and Zhou, Chenyu and Shen, Yunhang and Zhang, Mengdan and others},
  booktitle={Proceedings of the IEEE/CVF conference on computer vision and pattern recognition},
  pages={24108--24118},
  year={2025}
}

@inproceedings{petryk2024aloha,
  title={Aloha: A new measure for hallucination in captioning models},
  author={Petryk, Suzanne and Chan, David and Kachinthaya, Anish and Zou, Haodi and Canny, John and Gonzalez, Joseph and Darrell, Trevor},
  booktitle={Proceedings of the 2024 Conference of the North American Chapter of the Association for Computational Linguistics: Human Language Technologies (Volume 2: Short Papers)},
  pages={342--357},
  year={2024}
}

@inproceedings{chan2023clair,
  title={Clair: Evaluating image captions with large language models},
  author={Chan, David M and Petryk, Suzanne and Gonzalez, Joseph E and Darrell, Trevor and Canny, John},
  booktitle={Proceedings of the 2023 Conference on Empirical Methods in Natural Language Processing},
  pages={13638--13646},
  year={2023}
}

@article{lee2024toward,
  title={Toward robust hyper-detailed image captioning: A multiagent approach and dual evaluation metrics for factuality and coverage},
  author={Lee, Saehyung and Yoon, Seunghyun and Bui, Trung and Shi, Jing and Yoon, Sungroh},
  journal={arXiv preprint arXiv:2412.15484},
  year={2024}
}

@inproceedings{zhao2025mmvu,
  title={Mmvu: Measuring expert-level multi-discipline video understanding},
  author={Zhao, Yilun and Zhang, Haowei and Xie, Lujing and Hu, Tongyan and Gan, Guo and Long, Yitao and Hu, Zhiyuan and Chen, Weiyuan and Li, Chuhan and Xu, Zhijian and others},
  booktitle={Proceedings of the Computer Vision and Pattern Recognition Conference},
  pages={8475--8489},
  year={2025}
}

@inproceedings{li2024mvbench,
  title={Mvbench: A comprehensive multi-modal video understanding benchmark},
  author={Li, Kunchang and Wang, Yali and He, Yinan and Li, Yizhuo and Wang, Yi and Liu, Yi and Wang, Zun and Xu, Jilan and Chen, Guo and Luo, Ping and others},
  booktitle={Proceedings of the IEEE/CVF Conference on Computer Vision and Pattern Recognition},
  pages={22195--22206},
  year={2024}
}

@article{zhang2025timelens,
  title={Timelens: Rethinking video temporal grounding with multimodal llms},
  author={Zhang, Jun and Wang, Teng and Ge, Yuying and Ge, Yixiao and Li, Xinhao and Shan, Ying and Wang, Limin},
  journal={arXiv preprint arXiv:2512.14698},
  year={2025}
}

@inproceedings{gao2017tall,
  title={Tall: Temporal activity localization via language query},
  author={Gao, Jiyang and Sun, Chen and Yang, Zhenheng and Nevatia, Ram},
  booktitle={Proceedings of the IEEE international conference on computer vision},
  pages={5267--5275},
  year={2017}
}

@inproceedings{krishna2017dense,
  title={Dense-captioning events in videos},
  author={Krishna, Ranjay and Hata, Kenji and Ren, Frederic and Fei-Fei, Li and Carlos Niebles, Juan},
  booktitle={Proceedings of the IEEE international conference on computer vision},
  pages={706--715},
  year={2017}
}

@article{lei2021detecting,
  title={Detecting moments and highlights in videos via natural language queries},
  author={Lei, Jie and Berg, Tamara L and Bansal, Mohit},
  journal={Advances in Neural Information Processing Systems},
  volume={34},
  pages={11846--11858},
  year={2021}
}

@article{team2025kimi,
  title={Kimi {K}1. 5: Scaling reinforcement learning with llms},
  author={Team, Kimi and Du, Angang and Gao, Bofei and Xing, Bowei and Jiang, Changjiu and Chen, Cheng and Li, Cheng and Xiao, Chenjun and Du, Chenzhuang and Liao, Chonghua and others},
  journal={arXiv preprint arXiv:2501.12599},
  year={2025}
}

@article{chen2025avocado,
  title={Avocado: An audiovisual video captioner driven by temporal orchestration},
  author={Chen, Xinlong and Ding, Yue and Lin, Weihong and Hua, Jingyun and Yao, Linli and Shi, Yang and Li, Bozhou and Zhang, Yuanxing and Liu, Qiang and Wan, Pengfei and others},
  journal={arXiv preprint arXiv:2510.10395},
  year={2025}
}

@article{chu2025sft,
  title={{SFT} {M}emorizes, {RL} {G}eneralizes: A comparative study of foundation model post-training},
  author={Chu, Tianzhe and Zhai, Yuexiang and Yang, Jihan and Tong, Shengbang and Xie, Saining and Schuurmans, Dale and Le, Quoc V and Levine, Sergey and Ma, Yi},
  journal={arXiv preprint arXiv:2501.17161},
  year={2025}
}

@article{luo2025gui,
  title={GUI-R1: A generalist r1-style vision-language action model for gui agents},
  author={Luo, Run and Wang, Lu and He, Wanwei and Xia, Xiaobo},
  journal={arXiv preprint arXiv:2504.10458},
  year={2025}
}

@inproceedings{karpathy2015deep,
  title={Deep visual-semantic alignments for generating image descriptions},
  author={Karpathy, Andrej and Fei-Fei, Li},
  booktitle={Proceedings of the IEEE conference on computer vision and pattern recognition},
  pages={3128--3137},
  year={2015}
}

@inproceedings{vinyals2015show,
  title={Show and tell: A neural image caption generator},
  author={Vinyals, Oriol and Toshev, Alexander and Bengio, Samy and Erhan, Dumitru},
  booktitle={Proceedings of the IEEE conference on computer vision and pattern recognition},
  pages={3156--3164},
  year={2015}
}

@inproceedings{venugopalan2015sequence,
  title={Sequence to sequence-video to text},
  author={Venugopalan, Subhashini and Rohrbach, Marcus and Donahue, Jeffrey and Mooney, Raymond and Darrell, Trevor and Saenko, Kate},
  booktitle={Proceedings of the IEEE international conference on computer vision},
  pages={4534--4542},
  year={2015}
}

@inproceedings{donahue2015long,
  title={Long-term recurrent convolutional networks for visual recognition and description},
  author={Donahue, Jeffrey and Anne Hendricks, Lisa and Guadarrama, Sergio and Rohrbach, Marcus and Venugopalan, Subhashini and Saenko, Kate and Darrell, Trevor},
  booktitle={Proceedings of the IEEE conference on computer vision and pattern recognition},
  pages={2625--2634},
  year={2015}
}

@inproceedings{radford2021learning,
  title={Learning transferable visual models from natural language supervision},
  author={Radford, Alec and Kim, Jong Wook and Hallacy, Chris and Ramesh, Aditya and Goh, Gabriel and Agarwal, Sandhini and Sastry, Girish and Askell, Amanda and Mishkin, Pamela and Clark, Jack and others},
  booktitle={International conference on machine learning},
  pages={8748--8763},
  year={2021},
  organization={PmLR}
}

@article{liu2023visual,
  title={Visual instruction tuning},
  author={Liu, Haotian and Li, Chunyuan and Wu, Qingyang and Lee, Yong Jae},
  journal={Advances in neural information processing systems},
  volume={36},
  pages={34892--34916},
  year={2023}
}

@inproceedings{chen2024sharegpt4v,
  title={Sharegpt4v: Improving large multi-modal models with better captions},
  author={Chen, Lin and Li, Jinsong and Dong, Xiaoyi and Zhang, Pan and He, Conghui and Wang, Jiaqi and Zhao, Feng and Lin, Dahua},
  booktitle={European Conference on Computer Vision},
  pages={370--387},
  year={2024},
  organization={Springer}
}

@article{xing2025caprl,
  title={Caprl: Stimulating dense image caption capabilities via reinforcement learning},
  author={Xing, Long and Dong, Xiaoyi and Zang, Yuhang and Cao, Yuhang and Liang, Jianze and Huang, Qidong and Wang, Jiaqi and Wu, Feng and Lin, Dahua},
  journal={arXiv preprint arXiv:2509.22647},
  year={2025}
}

@inproceedings{masry2025chartqapro,
  title={Chartqapro: A more diverse and challenging benchmark for chart question answering},
  author={Masry, Ahmed and Islam, Mohammed Saidul and Ahmed, Mahir and Bajaj, Aayush and Kabir, Firoz and Kartha, Aaryaman and Laskar, Md Tahmid Rahman and Rahman, Mizanur and Rahman, Shadikur and Shahmohammadi, Mehrad and others},
  booktitle={Findings of the Association for Computational Linguistics: ACL 2025},
  pages={19123--19151},
  year={2025}
}

@inproceedings{qiao2025we,
  title={We-math: Does your large multimodal model achieve human-like mathematical reasoning?},
  author={Qiao, Runqi and Tan, Qiuna and Dong, Guanting and MinhuiWu, MinhuiWu and Sun, Chong and Song, Xiaoshuai and Wang, Jiapeng and Gongque, Zhuoma and Lei, Shanglin and Zhang, Yifan and others},
  booktitle={Proceedings of the 63rd Annual Meeting of the Association for Computational Linguistics (Volume 1: Long Papers)},
  pages={20023--20070},
  year={2025}
}

@inproceedings{yue2024mmmu,
  title={Mmmu: A massive multi-discipline multimodal understanding and reasoning benchmark for expert agi},
  author={Yue, Xiang and Ni, Yuansheng and Zhang, Kai and Zheng, Tianyu and Liu, Ruoqi and Zhang, Ge and Stevens, Samuel and Jiang, Dongfu and Ren, Weiming and Sun, Yuxuan and others},
  booktitle={Proceedings of the IEEE/CVF conference on computer vision and pattern recognition},
  pages={9556--9567},
  year={2024}
}

@article{li2023seed,
  title={Seed-bench: Benchmarking multimodal llms with generative comprehension},
  author={Li, Bohao and Wang, Rui and Wang, Guangzhi and Ge, Yuying and Ge, Yixiao and Shan, Ying},
  journal={arXiv preprint arXiv:2307.16125},
  year={2023}
}

@article{chen2024we,
  title={Are we on the right way for evaluating large vision-language models?},
  author={Chen, Lin and Li, Jinsong and Dong, Xiaoyi and Zhang, Pan and Zang, Yuhang and Chen, Zehui and Duan, Haodong and Wang, Jiaqi and Qiao, Yu and Lin, Dahua and others},
  journal={Advances in Neural Information Processing Systems},
  volume={37},
  pages={27056--27087},
  year={2024}
}

@article{wang2024measuring,
  title={Measuring multimodal mathematical reasoning with math-vision dataset},
  author={Wang, Ke and Pan, Junting and Shi, Weikang and Lu, Zimu and Ren, Houxing and Zhou, Aojun and Zhan, Mingjie and Li, Hongsheng},
  journal={Advances in Neural Information Processing Systems},
  volume={37},
  pages={95095--95169},
  year={2024}
}

@article{wang2024charxiv,
  title={Charxiv: Charting gaps in realistic chart understanding in multimodal llms},
  author={Wang, Zirui and Xia, Mengzhou and He, Luxi and Chen, Howard and Liu, Yitao and Zhu, Richard and Liang, Kaiqu and Wu, Xindi and Liu, Haotian and Malladi, Sadhika and others},
  journal={Advances in Neural Information Processing Systems},
  volume={37},
  pages={113569--113697},
  year={2024}
}

@inproceedings{zhang2024mathverse,
  title={Mathverse: Does your multi-modal llm truly see the diagrams in visual math problems?},
  author={Zhang, Renrui and Jiang, Dongzhi and Zhang, Yichi and Lin, Haokun and Guo, Ziyu and Qiu, Pengshuo and Zhou, Aojun and Lu, Pan and Chang, Kai-Wei and Qiao, Yu and others},
  booktitle={European Conference on Computer Vision},
  pages={169--186},
  year={2024},
  organization={Springer}
}

@inproceedings{yue2025mmmu,
  title={Mmmu-pro: A more robust multi-discipline multimodal understanding benchmark},
  author={Yue, Xiang and Zheng, Tianyu and Ni, Yuansheng and Wang, Yubo and Zhang, Kai and Tong, Shengbang and Sun, Yuxuan and Yu, Botao and Zhang, Ge and Sun, Huan and others},
  booktitle={Proceedings of the 63rd Annual Meeting of the Association for Computational Linguistics (Volume 1: Long Papers)},
  pages={15134--15186},
  year={2025}
}

@inproceedings{mathew2022infographicvqa,
  title={Infographicvqa},
  author={Mathew, Minesh and Bagal, Viraj and Tito, Rub{\`e}n and Karatzas, Dimosthenis and Valveny, Ernest and Jawahar, CV},
  booktitle={Proceedings of the IEEE/CVF Winter Conference on Applications of Computer Vision},
  pages={1697--1706},
  year={2022}
}

@article{abu2016youtube,
  title={Youtube-8m: A large-scale video classification benchmark},
  author={Abu-El-Haija, Sami and Kothari, Nisarg and Lee, Joonseok and Natsev, Paul and Toderici, George and Varadarajan, Balakrishnan and Vijayanarasimhan, Sudheendra},
  journal={arXiv preprint arXiv:1609.08675},
  year={2016}
}

@inproceedings{miech2019howto100m,
  title={Howto100m: Learning a text-video embedding by watching hundred million narrated video clips},
  author={Miech, Antoine and Zhukov, Dimitri and Alayrac, Jean-Baptiste and Tapaswi, Makarand and Laptev, Ivan and Sivic, Josef},
  booktitle={Proceedings of the IEEE/CVF international conference on computer vision},
  pages={2630--2640},
  year={2019}
}

@article{chen2024sharegpt4video,
  title={Sharegpt4video: Improving video understanding and generation with better captions},
  author={Chen, Lin and Wei, Xilin and Li, Jinsong and Dong, Xiaoyi and Zhang, Pan and Zang, Yuhang and Chen, Zehui and Duan, Haodong and Lin, Bin and Tang, Zhenyu and others},
  journal={Advances in Neural Information Processing Systems},
  volume={37},
  pages={19472--19495},
  year={2024}
}

@article{clark2026molmo2,
  title={Molmo2: Open Weights and Data for Vision-Language Models with Video Understanding and Grounding},
  author={Clark, Christopher and Zhang, Jieyu and Ma, Zixian and Park, Jae Sung and Salehi, Mohammadreza and Tripathi, Rohun and Lee, Sangho and Ren, Zhongzheng and Kim, Chris Dongjoo and Yang, Yinuo and others},
  journal={arXiv preprint arXiv:2601.10611},
  year={2026}
}

@inproceedings{yang2023vid2seq,
  title={Vid2seq: Large-scale pretraining of a visual language model for dense video captioning},
  author={Yang, Antoine and Nagrani, Arsha and Seo, Paul Hongsuck and Miech, Antoine and Pont-Tuset, Jordi and Laptev, Ivan and Sivic, Josef and Schmid, Cordelia},
  booktitle={Proceedings of the IEEE/CVF conference on computer vision and pattern recognition},
  pages={10714--10726},
  year={2023}
}

@inproceedings{maaz2024video,
  title={Video-chatgpt: Towards detailed video understanding via large vision and language models},
  author={Maaz, Muhammad and Rasheed, Hanoona and Khan, Salman and Khan, Fahad},
  booktitle={Proceedings of the 62nd Annual Meeting of the Association for Computational Linguistics (Volume 1: Long Papers)},
  pages={12585--12602},
  year={2024}
}

@article{wang2022internvideo,
  title={Internvideo: General video foundation models via generative and discriminative learning},
  author={Wang, Yi and Li, Kunchang and Li, Yizhuo and He, Yinan and Huang, Bingkun and Zhao, Zhiyu and Zhang, Hongjie and Xu, Jilan and Liu, Yi and Wang, Zun and others},
  journal={arXiv preprint arXiv:2212.03191},
  year={2022}
}

@inproceedings{bain2021frozen,
  title={Frozen in time: A joint video and image encoder for end-to-end retrieval},
  author={Bain, Max and Nagrani, Arsha and Varol, G{\"u}l and Zisserman, Andrew},
  booktitle={Proceedings of the IEEE/CVF international conference on computer vision},
  pages={1728--1738},
  year={2021}
}

@inproceedings{rotstein2024fusecap,
  title={Fusecap: Leveraging large language models for enriched fused image captions},
  author={Rotstein, Noam and Bensaid, David and Brody, Shaked and Ganz, Roy and Kimmel, Ron},
  booktitle={Proceedings of the IEEE/CVF winter conference on applications of computer vision},
  pages={5689--5700},
  year={2024}
}

@inproceedings{vasu2025fastvlm,
  title={Fastvlm: Efficient vision encoding for vision language models},
  author={Vasu, Pavan Kumar Anasosalu and Faghri, Fartash and Li, Chun-Liang and Koc, Cem and True, Nate and Antony, Albert and Santhanam, Gokula and Gabriel, James and Grasch, Peter and Tuzel, Oncel and others},
  booktitle={Proceedings of the Computer Vision and Pattern Recognition Conference},
  pages={19769--19780},
  year={2025}
}

@article{liu2025inference,
  title={Inference-time scaling for generalist reward modeling},
  author={Liu, Zijun and Wang, Peiyi and Xu, Runxin and Ma, Shirong and Ruan, Chong and Li, Peng and Liu, Yang and Wu, Yu},
  journal={arXiv preprint arXiv:2504.02495},
  year={2025}
}

@article{su2025crossing,
  title={Crossing the Reward Bridge: Expanding RL with Verifiable Rewards Across Diverse Domains},
  author={Su, Yi and Yu, Dian and Song, Linfeng and Li, Juntao and Mi, Haitao and Tu, Zhaopeng and Zhang, Min and Yu, Dong},
  journal={arXiv preprint arXiv:2503.23829},
  year={2025}
}

@article{gunjal2025rubrics,
  title={Rubrics as rewards: Reinforcement learning beyond verifiable domains},
  author={Gunjal, Anisha and Wang, Anthony and Lau, Elaine and Nath, Vaskar and Liu, Bing and Hendryx, Sean},
  journal={arXiv preprint arXiv:2507.17746},
  year={2025}
}

@article{lu2025writing,
  title={Writing-Zero: Bridge the Gap Between Non-verifiable Problems and Verifiable Rewards},
  author={Lu, Xun},
  journal={arXiv preprint arXiv:2506.00103},
  year={2025}
}

@article{gurung2025learning,
  title={Learning to reason for long-form story generation},
  author={Gurung, Alexander and Lapata, Mirella},
  journal={arXiv preprint arXiv:2503.22828},
  year={2025}
}

@article{yu2025rlpr,
  title={{RLPR}: Extrapolating RLVR to General Domains without Verifiers},
  author={Yu, Tianyu and Ji, Bo and Wang, Shouli and Yao, Shu and Wang, Zefan and Cui, Ganqu and Yuan, Lifan and Ding, Ning and Yao, Yuan and Liu, Zhiyuan and others},
  journal={arXiv preprint arXiv:2506.18254},
  year={2025}
}

@article{schuhmann2022laion,
  title={Laion-5b: An open large-scale dataset for training next generation image-text models},
  author={Schuhmann, Christoph and Beaumont, Romain and Vencu, Richard and Gordon, Cade and Wightman, Ross and Cherti, Mehdi and Coombes, Theo and Katta, Aarush and Mullis, Clayton and Wortsman, Mitchell and others},
  journal={Advances in neural information processing systems},
  volume={35},
  pages={25278--25294},
  year={2022}
}

@inproceedings{changpinyo2021conceptual,
  title={Conceptual 12m: Pushing web-scale image-text pre-training to recognize long-tail visual concepts},
  author={Changpinyo, Soravit and Sharma, Piyush and Ding, Nan and Soricut, Radu},
  booktitle={Proceedings of the IEEE/CVF conference on computer vision and pattern recognition},
  pages={3558--3568},
  year={2021}
}

@article{thomee2016yfcc100m,
  title={Yfcc100m: The new data in multimedia research},
  author={Thomee, Bart and Shamma, David A and Friedland, Gerald and Elizalde, Benjamin and Ni, Karl and Poland, Douglas and Borth, Damian and Li, Li-Jia},
  journal={Communications of the ACM},
  volume={59},
  number={2},
  pages={64--73},
  year={2016},
  publisher={ACM New York, NY, USA}
}

@inproceedings{li2022blip,
  title={Blip: Bootstrapping language-image pre-training for unified vision-language understanding and generation},
  author={Li, Junnan and Li, Dongxu and Xiong, Caiming and Hoi, Steven},
  booktitle={International Conference on Machine Learning},
  pages={12888--12900},
  year={2022},
  organization={PMLR}
}

@article{fan2023improving,
  title={Improving clip training with language rewrites},
  author={Fan, Lijie and Krishnan, Dilip and Isola, Phillip and Katabi, Dina and Tian, Yonglong},
  journal={Advances in Neural Information Processing Systems},
  volume={36},
  pages={35544--35575},
  year={2023}
}

@inproceedings{yu2024capsfusion,
  title={Capsfusion: Rethinking image-text data at scale},
  author={Yu, Qiying and Sun, Quan and Zhang, Xiaosong and Cui, Yufeng and Zhang, Fan and Cao, Yue and Wang, Xinlong and Liu, Jingjing},
  booktitle={Proceedings of the IEEE/CVF Conference on Computer Vision and Pattern Recognition},
  pages={14022--14032},
  year={2024}
}

@article{chen2024allava,
  title={Allava: Harnessing gpt4v-synthesized data for lite vision-language models},
  author={Chen, Guiming Hardy and Chen, Shunian and Zhang, Ruifei and Chen, Junying and Wu, Xiangbo and Zhang, Zhiyi and Chen, Zhihong and Li, Jianquan and Wan, Xiang and Wang, Benyou},
  journal={arXiv preprint arXiv:2402.11684},
  year={2024}
}

@article{li2024densefusion,
  title={Densefusion-1m: Merging vision experts for comprehensive multimodal perception},
  author={Li, Xiaotong and Zhang, Fan and Diao, Haiwen and Wang, Yueze and Wang, Xinlong and Duan, Ling-Yu},
  journal={arXiv preprint arXiv:2407.08303},
  year={2024}
}

@article{wang2025unified,
  title={Unified reward model for multimodal understanding and generation},
  author={Wang, Yibin and Zang, Yuhang and Li, Hao and Jin, Cheng and Wang, Jiaqi},
  journal={arXiv preprint arXiv:2503.05236},
  year={2025}
}

@article{sun2024descriptive,
  title={Descriptive Caption Enhancement with Visual Specialists for Multimodal Perception},
  author={Sun, Yanpeng and Hao, Jing and Zhu, Ke and Liu, Jiang-Jiang and Zhao, Yuxiang and Li, Xiaofan and Zhang, Gang and Li, Zechao and Wang, Jingdong},
  journal={arXiv preprint arXiv:2412.14233},
  year={2024}
}

@article{qiao2024prism,
  title={Prism: A framework for decoupling and assessing the capabilities of vlms},
  author={Qiao, Yuxuan and Duan, Haodong and Fang, Xinyu and Yang, Junming and Chen, Lin and Zhang, Songyang and Wang, Jiaqi and Lin, Dahua and Chen, Kai},
  journal={Advances in Neural Information Processing Systems},
  volume={37},
  pages={111863--111898},
  year={2024}
}

@article{guo2025deepseek,
  title={{Deepseek-R1}: Incentivizing reasoning capability in llms via reinforcement learning},
  author={Guo, Daya and Yang, Dejian and Zhang, Haowei and Song, Junxiao and Zhang, Ruoyu and Xu, Runxin and Zhu, Qihao and Ma, Shirong and Wang, Peiyi and Bi, Xiao and others},
  journal={arXiv preprint arXiv:2501.12948},
  year={2025}
}

@misc{flux2024,
    author={Black Forest Labs},
    title={FLUX},
    year={2024},
    howpublished={\url{https://github.com/black-forest-labs/flux}},
}

@misc{flux-2-2025,
    author={Black Forest Labs},
    title={{FLUX.2: Frontier Visual Intelligence}},
    year={2025},
    howpublished={\url{https://bfl.ai/blog/flux-2}},
}

@article{chen2024open,
  title={Open-llava-next: An open-source implementation of llava-next series for facilitating the large multi-modal model community},
  author={Chen, Lin and Xing, Long},
  journal={GitHub-xiaoachen98/Open-LLaVA-NeXT: Anopen-sourceimplementationfortrainingLLaVA-NeXT},
  year={2024}
}

@article{li2025otter,
  title={Otter: A multi-modal model with in-context instruction tuning},
  author={Li, Bo and Zhang, Yuanhan and Chen, Liangyu and Wang, Jinghao and Pu, Fanyi and Cahyono, Joshua Adrian and Yang, Jingkang and Li, Chunyuan and Liu, Ziwei},
  journal={IEEE Transactions on Pattern Analysis and Machine Intelligence},
  year={2025},
  publisher={IEEE}
}

@article{abdar2024review,
  title={A review of deep learning for video captioning},
  author={Abdar, Moloud and Kollati, Meenakshi and Kuraparthi, Swaraja and Pourpanah, Farhad and McDuff, Daniel and Ghavamzadeh, Mohammad and Yan, Shuicheng and Mohamed, Abduallah and Khosravi, Abbas and Cambria, Erik and others},
  journal={IEEE Transactions on Pattern Analysis and Machine Intelligence},
  year={2024},
  publisher={IEEE}
}

@article{li2025uni,
  title={Uni-moe: Scaling unified multimodal llms with mixture of experts},
  author={Li, Yunxin and Jiang, Shenyuan and Hu, Baotian and Wang, Longyue and Zhong, Wanqi and Luo, Wenhan and Ma, Lin and Zhang, Min},
  journal={IEEE Transactions on Pattern Analysis and Machine Intelligence},
  year={2025},
  publisher={IEEE}
}

@article{wang2025video,
  title={Video dataflywheel: Resolving the impossible data trinity in video-language understanding},
  author={Wang, Xiao and Wu, Jianlong and Lin, Zijia and Zhang, Fuzheng and Zhang, Di and Nie, Liqiang},
  journal={IEEE Transactions on Pattern Analysis and Machine Intelligence},
  year={2025},
  publisher={IEEE}
}

@article{stefanini2022show,
  title={From show to tell: A survey on deep learning-based image captioning},
  author={Stefanini, Matteo and Cornia, Marcella and Baraldi, Lorenzo and Cascianelli, Silvia and Fiameni, Giuseppe and Cucchiara, Rita},
  journal={IEEE transactions on pattern analysis and machine intelligence},
  volume={45},
  number={1},
  pages={539--559},
  year={2022},
  publisher={IEEE}
}

@article{zhang2024vision,
  title={Vision-language models for vision tasks: A survey},
  author={Zhang, Jingyi and Huang, Jiaxing and Jin, Sheng and Lu, Shijian},
  journal={IEEE transactions on pattern analysis and machine intelligence},
  volume={46},
  number={8},
  pages={5625--5644},
  year={2024},
  publisher={IEEE}
}

@article{vinyals2016show,
  title={Show and tell: Lessons learned from the 2015 mscoco image captioning challenge},
  author={Vinyals, Oriol and Toshev, Alexander and Bengio, Samy and Erhan, Dumitru},
  journal={IEEE transactions on pattern analysis and machine intelligence},
  volume={39},
  number={4},
  pages={652--663},
  year={2016},
  publisher={IEEE}
}

@article{ma2024robust,
  title={Robust visual question answering: Datasets, methods, and future challenges},
  author={Ma, Jie and Wang, Pinghui and Kong, Dechen and Wang, Zewei and Liu, Jun and Pei, Hongbin and Zhao, Junzhou},
  journal={IEEE Transactions on Pattern Analysis and Machine Intelligence},
  volume={46},
  number={8},
  pages={5575--5594},
  year={2024},
  publisher={IEEE}
}

@article{xu2023multimodal,
  title={Multimodal learning with transformers: A survey},
  author={Xu, Peng and Zhu, Xiatian and Clifton, David A},
  journal={IEEE Transactions on Pattern Analysis and Machine Intelligence},
  volume={45},
  number={10},
  pages={12113--12132},
  year={2023},
  publisher={Ieee}
}

@article{wu2025survey,
  title={A survey on video temporal grounding with multimodal large language model},
  author={Wu, Jianlong and Liu, Wei and Liu, Ye and Liu, Meng and Nie, Liqiang and Lin, Zhouchen and Chen, Chang Wen},
  journal={IEEE Transactions on Pattern Analysis and Machine Intelligence},
  year={2025},
  publisher={IEEE}
}

@article{wang2020diversity,
  title={On diversity in image captioning: Metrics and methods},
  author={Wang, Qingzhong and Wan, Jia and Chan, Antoni B},
  journal={IEEE Transactions on Pattern Analysis and Machine Intelligence},
  volume={44},
  number={2},
  pages={1035--1049},
  year={2020},
  publisher={IEEE}
}

@inproceedings{lin2004rouge,
  title={Rouge: A package for automatic evaluation of summaries},
  author={Lin, Chin-Yew},
  booktitle={Text summarization branches out},
  pages={74--81},
  year={2004}
}

@inproceedings{papineni2002bleu,
  title={Bleu: a method for automatic evaluation of machine translation},
  author={Papineni, Kishore and Roukos, Salim and Ward, Todd and Zhu, Wei-Jing},
  booktitle={Proceedings of the 40th annual meeting of the Association for Computational Linguistics},
  pages={311--318},
  year={2002}
}

@article{abbas2023semdedup,
  title={Semdedup: Data-efficient learning at web-scale through semantic deduplication},
  author={Abbas, Amro and Tirumala, Kushal and Simig, D{\'a}niel and Ganguli, Surya and Morcos, Ari S},
  journal={arXiv preprint arXiv:2303.09540},
  year={2023}
}

@article{liu2025visual,
  title={Visual-RFT: Visual Reinforcement Fine-Tuning},
  author={Liu, Ziyu and Sun, Zeyi and Zang, Yuhang and Dong, Xiaoyi and Cao, Yuhang and Duan, Haodong and Lin, Dahua and Wang, Jiaqi},
  journal={arXiv preprint arXiv:2503.01785},
  year={2025}
}

@article{xue2025dancegrpo,
  title={DanceGRPO: Unleashing GRPO on Visual Generation},
  author={Xue, Zeyue and Wu, Jie and Gao, Yu and Kong, Fangyuan and Zhu, Lingting and Chen, Mengzhao and Liu, Zhiheng and Liu, Wei and Guo, Qiushan and Huang, Weilin and others},
  journal={arXiv preprint arXiv:2505.07818},
  year={2025}
}

@article{Pref-GRPO&UniGenBench,
  title={Pref-GRPO: Pairwise Preference Reward-based GRPO for Stable Text-to-Image Reinforcement Learning},
  author={Wang, Yibin and Li, Zhimin and Zang, Yuhang and Zhou, Yujie and Bu, Jiazi and Wang, Chunyu and Lu, Qinglin and Jin, Cheng and Wang, Jiaqi},
  journal={arXiv preprint arXiv:2508.20751},
  year={2025}
}

\end{document}